\pdfoutput=1

\documentclass[11pt]{article}

\usepackage[preprint]{acl}

\usepackage{times}
\usepackage{latexsym}
\usepackage{url}

\usepackage[T1]{fontenc}

\usepackage[utf8]{inputenc}
\usepackage{CJKutf8}
\usepackage{microtype}

\usepackage{inconsolata}

\usepackage{float}
\usepackage{booktabs}
\usepackage{enumitem}
\usepackage{graphicx}
\usepackage{latexsym}
\usepackage{color}
\usepackage{makecell}
\usepackage{multirow}
\usepackage{booktabs}
\usepackage{subfigure}
\usepackage{multicol}
\usepackage{adjustbox}

\usepackage{tcolorbox}  
\usepackage{geometry}   
\usepackage{xcolor}
\usepackage{tcolorbox}
\definecolor{translationcolor}{RGB}{0,102,204} 
\definecolor{correctcolor}{RGB}{0,153,0} 

\usepackage{graphicx}
\usepackage{tikz}
\usetikzlibrary{positioning}

\title{Fùxì: A Benchmark for Evaluating Language Models on Ancient Chinese Text Understanding and Generation}

\author{
    \textbf{Shangqing Zhao\textsuperscript{$1$}},
    \textbf{Yuhao Zhou\textsuperscript{$1$}},
    \textbf{Yupei Ren\textsuperscript{$1$,$2$,$3$}}, \\ 
    \textbf{Zhe Chen\textsuperscript{$1$}},
    \textbf{Chenghao Jia\textsuperscript{$1$}},
    \textbf{Fang Zhe\textsuperscript{$1$}},
    \textbf{Zhaogaung Long\textsuperscript{$1$}},
    \textbf{Shu Liu\textsuperscript{$1$,$2$,$3$}},
    \textbf{Man Lan\textsuperscript{$1$,$2$,$3$}} \\
    \textsuperscript{1}School of Computer Science and Technology, East China Normal University \\
    \textsuperscript{2}Lab of Artificial Intelligence for Education, East China Normal University \\
    \textsuperscript{3}Shanghai Institute of Artificial Intelligence for Education, East China Normal University\\
    \texttt{sqzhao@stu.ecnu.edu.cn, mlan@cs.ecnu.edu.cn} \\
}

\usepackage{subfigure}
\usepackage{pgfplots}
\pgfplotsset{compat=1.18}
\usepackage{graphicx}

\begin{document}
\maketitle
\begin{CJK}{UTF8}{gbsn}   
\begin{abstract}

Ancient Chinese text processing presents unique challenges for large language models (LLMs) due to its distinct linguistic features, complex structural constraints, and rich cultural context. While existing benchmarks have primarily focused on evaluating comprehension through multiple-choice questions, there remains a critical gap in assessing models' generative capabilities in classical Chinese. We introduce Fùxì, a comprehensive benchmark that evaluates both understanding and generation capabilities across 21 diverse tasks. Our benchmark distinguishes itself through three key contributions: (1) balanced coverage of both comprehension and generation tasks, including novel tasks like poetry composition and couplet completion, (2) specialized evaluation metrics designed specifically for classical Chinese text generation, combining rule-based verification with fine-tuned LLM evaluators, and (3) a systematic assessment framework that considers both linguistic accuracy and cultural authenticity. Through extensive evaluation of state-of-the-art LLMs, we reveal significant performance gaps between understanding and generation tasks, with models achieving promising results in comprehension but struggling considerably in generation tasks, particularly those requiring deep cultural knowledge and adherence to classical formats. Our findings highlight the current limitations in ancient Chinese text processing and provide insights for future model development. The benchmark, evaluation toolkit, and baseline results are publicly available to facilitate research in this domain\footnote{\url{https://github.com/cubenlp/FuxiBench}}.

\end{abstract}

\section{Introduction}

Ancient Chinese, as a written form of the Chinese language used for more than two millennia in East Asia, has profoundly shaped the cultural and intellectual development of the region~\cite{holcombe2006rethinking, SYSJ200903026}. With an estimated corpus of 400 million words spanning 3 million texts, this vast literary heritage encompasses literature, philosophy, history, and various other domains, with approximately half of these texts holding significant academic and cultural value~\cite{DXTS200201005,zhouWYWEBNLPEvaluation2023a, XDXK202413029}.

\begin{figure}[t]
    \centering
        \resizebox{\hsize}{!}{
            \includegraphics[scale=0.5]{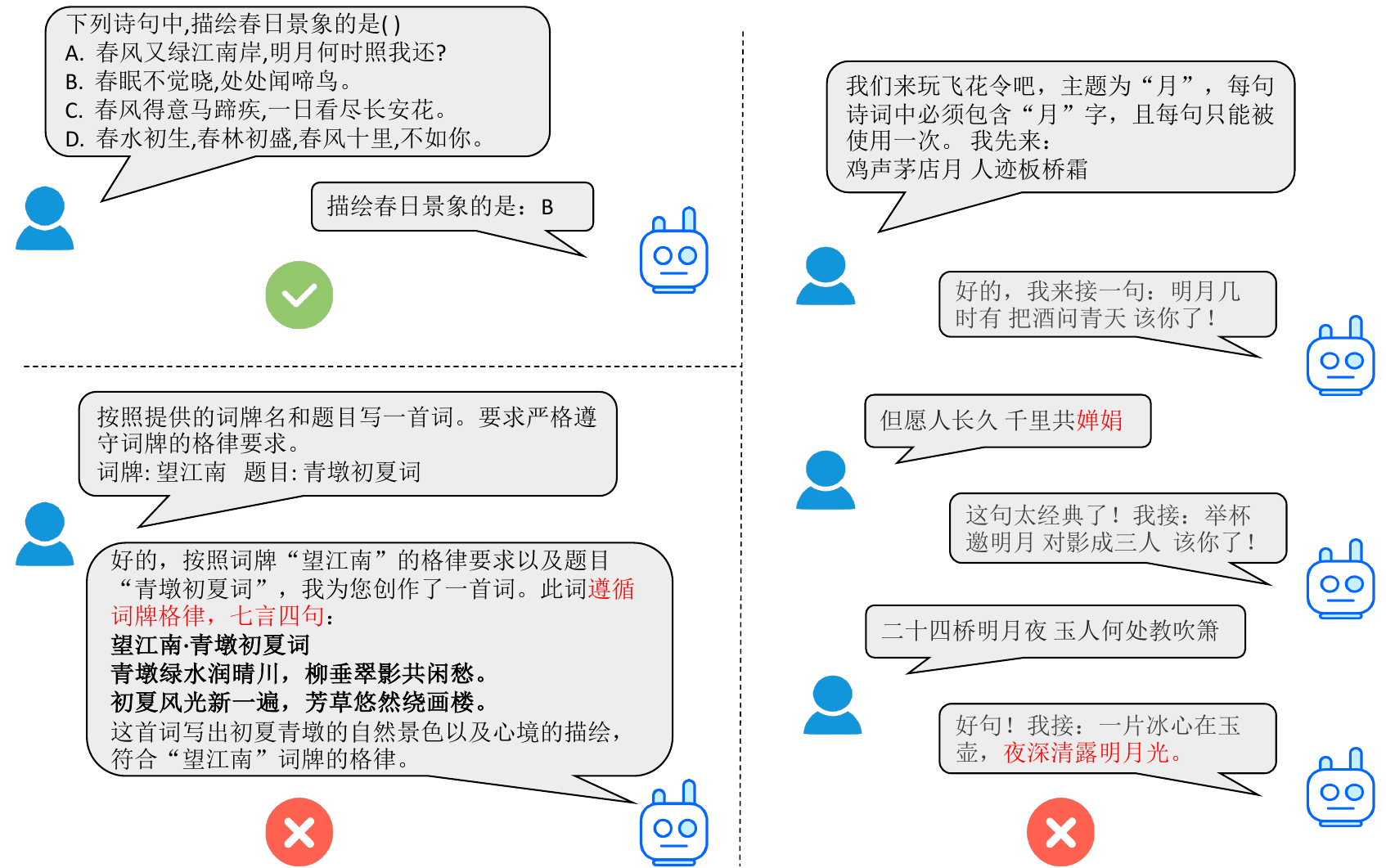}
        }
        \caption{ Interactions with GPT-4 about Ancient Chinese related topics. }
        \label{fig:fig1}
\end{figure}

However, the increasing gap between classical and modern Chinese poses significant challenges for contemporary readers and researchers. This linguistic evolution necessitates advanced Natural Language Processing (NLP) technologies to effectively process, understand, and analyze classical Chinese texts. While pre-trained language models like BERT~\cite{devlinBERTPretrainingDeep2019} and GPT~\cite{radfordImprovingLanguageUnderstanding2018} have demonstrated remarkable capabilities in modern Chinese NLP tasks, their application to classical Chinese presents unique challenges due to substantial differences in grammar, vocabulary, and writing conventions. As illustrated in Figure~\ref{fig:fig1}, even GPT-4~\cite{openai2023GPT4TechnicalReport}, currently one of the most advanced LLMs, struggles with traditional Chinese poetry games. Although these models can sometimes correctly answer multiple-choice questions by selecting from given options, their performance in generating Ancient Chinese related text still needs improvement, highlighting the complexity of classical Chinese language understanding and generation.

While existing benchmarks such as C-CLUE~\cite{jiCCLUEBenchmarkClassical2021} and WYWEB~\cite{zhouWYWEBNLPEvaluation2023a} have made important contributions in evaluating model performance on specific tasks, and more recent works like ACLUE~\cite{zhang2023CanLargeLangauge} and AC-EVAL~\cite{wei2024ACEVALEvaluatingAncient} have provided comprehensive evaluation frameworks, there remains a critical gap in assessing models' generative capabilities in classical Chinese. 
Most benchmarks rely heavily on multiple-choice questions, which are useful for assessing comprehension but do not effectively measure a model’s ability to generate coherent and contextually appropriate classical Chinese text.

To address these limitations, we introduce Fùxì (负\adjustbox{valign=m}{\includegraphics[height=1em]{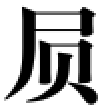}}), a novel benchmark designed to comprehensively evaluate both the understanding and generation capabilities of language models in classical Chinese. Fùxì distinguishes itself through:
    1) A balanced focus on both comprehension and generation tasks, with 21 carefully designed task types spanning various aspects of classical Chinese;
    2) Novel evaluation metrics specifically developed for assessing the quality of generated classical Chinese text;
    3) A comprehensive evaluation framework that considers both linguistic accuracy and cultural authenticity.

Our contributions are summarized as follows:
\begin{itemize}
    \item We present Fùxì, the first benchmark to comprehensively evaluate both understanding and generation capabilities in classical Chinese, featuring 21 diverse task types and novel evaluation metrics
    
    \item We introduce innovative task types specifically designed to assess unique aspects of classical Chinese, including poetry generation, couplet completion, and style transfer between classical and modern Chinese
    
    \item We provide extensive baseline results for state-of-the-art language models on Fùxì, revealing critical insights about their strengths and limitations in handling classical Chinese
    
    \item We release our benchmark, evaluation toolkit, and a fine-tuned model specialized for classical Chinese text evaluation, facilitating further research in this domain
\end{itemize}

Through Fùxì, we aim to advance the development of more capable language models for processing classical Chinese texts, ultimately contributing to the preservation and accessibility of this rich cultural heritage.

\section{Related Work}
\subsection{Chinese Benchmarks for LMs}
Recent years have witnessed significant progress in developing comprehensive benchmarks to evaluate Chinese language models. Early benchmarks such as MMCU~\cite{zeng2023MeasuringMassiveMultitask} and C-Eval~\cite{huang2023CEvalMultiLevelMultiDisciplinea} focused on assessing models' performance across diverse disciplines through multiple-choice questions derived from standardized tests. These were followed by more sophisticated frameworks like CMMLU~\cite{li2024CMMLUMeasuringMassive} and SuperCLUE~\cite{xu2023SuperCLUEComprehensiveChinese}, which incorporated both closed- and open-ended tasks to evaluate reasoning capabilities and linguistic understanding.

Recognizing the importance of text generation, benchmarks like CG-Eval~\cite{zeng2024EvaluatingGenerationCapabilities} and CLEVA~\cite{li2023CLEVAChineseLanguage} introduced tasks specifically designed to evaluate generative capabilities, including short answer generation, computational problem-solving, and creative writing tasks. OpenCompass~\cite{2023opencompass}, integrating over 100 datasets, further expanded the scope of generation evaluation. While these benchmarks have significantly advanced Chinese language model evaluation, they primarily focus on modern Chinese, leaving the unique challenges of classical Chinese largely unexplored.

\subsection{Ancient Chinese Benchmarks}
Ancient Chinese, with its rich linguistic and cultural heritage, necessitates specialized resources and evaluation tools to assess its historical, literary, and syntactic complexities. Current evaluation benchmarks can be categorized into three main streams: understanding-focused, comprehensive evaluation frameworks, and generation-oriented assessments.

For specific understanding tasks, several benchmarks have emerged: GuNER~\cite{su2023CCL23EvalRenWu1ZongJieBaoGaoGuJiMingMingShiTiShiBieGuNER2023OverviewCCL23Eval} for named entity recognition, GuwenEE\footnote{https://github.com/Lyn4ever29/GuwenEE} for event extraction, and CCLUE~\cite{jiCCLUEBenchmarkClassical2021} for general language understanding. Complementary resources such as CCP-KG~\cite{yutong2020ConstructionAnalysisClassical} and CQY~\cite{shao2021SentimentStyleControllable} address specialized aspects such as classical poetry sentiment analysis. For cross-temporal understanding, translation-focused datasets such as Erya~\cite{guo2023EffectiveAncientChinese} and EvaHan~\cite{wang-etal-2023-evahan2023} specifically address the syntactic and semantic relationships between ancient and modern Chinese.

Recent developments have introduced more comprehensive evaluation frameworks. ACHNLU~\cite{hu2024EvaluationAncientChinese} evaluated 12 mainstream LLMs' natural language understanding capabilities, while ACLUE~\cite{zhang2023CanLargeLangauge} presented a framework encompassing 15 tasks across phonetic, lexical, syntactic, semantic, inference and knowledge-based dimensions. Based on these, AC-EVAL introduced a three-tier framework to assess general historical knowledge, short text understanding, and long text comprehension through 13 distinct tasks. However, these frameworks predominantly rely on multiple-choice questions, which, while facilitating standardized evaluation, may not fully capture the models' tendency toward hallucination or their generative capabilities.

In terms of classical text generation, systems like Jiuge~\cite{li2021PoeticExpressionScenery} and Yu Sheng~\cite{ma2023YuShengHumaninLoop} have demonstrated the potential for the generation of ancient Chinese poetry, although their evaluation is based primarily on human evaluation. A notable advancement is CharPoet~\cite{yu2024CharPoetChineseClassical}, which focuses on improving format correctness through token-free generation pruning of LLMs.

Although existing benchmarks have made significant strides in evaluating ancient Chinese language understanding, they either lack comprehensive coverage or rely heavily on multiple-choice assessments. Notably, the evaluation of LLMs' capability to generate authentic classical Chinese content remains largely unexplored and requires systematic investigation.

\subsection{Evaluation Metrics for LLMs}
The evaluation of language models, particularly for classical Chinese text generation, presents unique challenges that require specialized metrics and approaches. Traditional metrics like BLEU~\cite{papineni2002bleu}, ROUGE~\cite{lin2004rouge}, and METEOR~\cite{banerjee2005meteor} have been widely used for assessing text generation quality. However, these metrics often fall short in capturing the nuances of classical Chinese, especially its unique structural requirements, linguistic patterns, and rich cultural elements that are essential for authentic classical text generation.

Human evaluation represents another approach, where human judges assess responses generated by different LLMs based on criteria such as accuracy, fluency, and cultural authenticity~\cite{Wei2024SystematicEO}. While human evaluation provides nuanced assessment, it is prohibitively slow and costly, making it impractical for large-scale evaluations and rapid development cycles. To address these limitations, researchers have proposed using commercial LLMs as surrogate evaluators, known as LLM-as-a-judge~\cite{Wei2024SystematicEO}, where models like GPT-4 and GPT-3.5-turbo serve as automated judges. However, these LLM judges often exhibit biases and inconsistencies in their evaluations, even when various bias-mitigation techniques are applied.
In this work, we adopt a hybrid approach combining rule-based metrics with aligned LLM evaluators. For tasks with well-defined structural constraints, we implement rule-based verification to assess adherence to classical Chinese literary forms. For more subjective aspects requiring deeper linguistic and cultural understanding, we employ fine-tuned LLMs specifically aligned to evaluate classical Chinese text generation, mitigating the biases present in general-purpose commercial LLMs while maintaining evaluation efficiency.

\section{Fùxì Overview}

\begin{table*}[]
    \resizebox{\textwidth}{!}{%

        \begin{tabular}{@{}llcc@{}}
        \toprule
        \textbf{Task}                    & \textbf{Description}                                          & \textbf{Type}                & \textbf{Eval Metric}  \\ \midrule
        Ancient Chinese RC               & Multiple choice question on ancient Chinese text reading comprehension & NLU                 & Acc \\
        Idiom RC                         & Multiple choice question on idiom reading comprehension                & NLU                 & Acc \\
        TCM Syndrome RC                  & Multiple choice question on TCM syndrome differentiation               & NLU                 & Acc \\ \hline 
        Loan Character QA                & Loan character identification question answering                       & NLG                 & Acc \\
        Allegorical Saying QA            & Allegorical saying question answering, to complete the second half.    & NLG                 & Acc \\
        Book Author                      & Given a book title, answer the author                                  & NLG                 & Acc \\
        Book Dynasty                     & Given a book title, answer the dynasty                                 & NLG                 & Acc \\
        Book Collection Classification   & Classify book titles into 10 collections                               & NLG                 & Acc \\
        Poetry Generation                & Poetry creation according to topic words                               & NLG                 & Acc \\
        Poetry Line Source Tracing       & Identify the source of a poem line                                     & NLG                 & Acc \\
        Famous Quote Source Tracing      & Identify the source of a classical Chinese quote                       & NLG                 & Acc \\
        Idiom Source Tracing             & Name the source book or essay of an idiom.                             & NLG                 & Acc \\ 
        Inverse Poetry Translation       & Decipher the original poem from translated modern Chinese version.     & NLG                 & Acc \\ \hline 
        Poetry Translation               & Translate a poem into modern Chinese                                   & NLG                 & BLEU \\ 
        Ancient Chinese Translation      & Translate ancient Chinese text into modern Chinese(extracted from the erya dataset) & NLG    & BLEU \\ 
        Poetry Appreciation              & Analysis of imagery, style, sentiment in classical Chinese poetry      & NLG                 & BLEU \\
        Idiom Explanation                & Idiom explanation                                                      & NLG                 & BLEU \\ 
        TCM QA                           & Answer questions given related TCM text.                               & NLG                 & BLEU \\
        Prescription Explanation         & Prescription details for TCM                                           & NLG                 & BLEU \\ \hline 
        Couplet Generation               & Generate the second line based on the first line.                      & NLG                 & Acc \\
        Ci Generation                    & Write a poem following a specific ci pattern with tonal rules.         & NLG                 & Acc \\
        \bottomrule
        \end{tabular}%
    }
    \caption{Overview of the Fuxi benchmark, including task description and evaluation metrics.}
    \label{tab:overview}
\end{table*}

\subsection{Design Principles}

\noindent \textbf{Task Diversity}
Previous work has predominantly focused on assessing comprehension abilities through multiple-choice questions. However, text generation is equally critical for ancient Chinese language processing. In the Fùxì benchmark, we emphasize the evaluation of models' generative capabilities by introducing a variety of generation tasks. These include tasks such as classical-to-modern Chinese translation, poetry appreciation, and completing missing lines in poetic compositions.

\noindent \textbf{Data Quality}
Our dataset is derived from high-quality internet sources, with manual curation to consolidate duplicate content and ensure data consistency. For deterministic tasks, such as predicting the subsequent line of a poem, we automated the data construction process using custom scripts to maintain reliability. Additionally, some tasks are sourced from pre-annotated datasets, further ensuring the quality and integrity of the task data.

\noindent \textbf{Task Evaluability}
Unlike prior benchmarks that primarily rely on multiple-choice formats for ease of scoring, Fùxì includes a significant number of tasks that require text generation as answers. This introduces unique challenges in evaluation. For tasks without strictly correct answers, such as classical-to-modern Chinese translation and poetry appreciation, we employ widely-used text generation evaluation metrics like BLEU~\cite{papineni2002bleu}. For tasks with definitive answers, assessing the correctness of model-generated text presents a significant challenge. While human evaluation is an option, it is resource-intensive and subject to inter-annotator variability. Additionally, there is a lack of established automated metrics for such tasks. To enable consistent and automated evaluation, we developed a prompt-based framework and manually annotated 1,000 samples, which were used to fine-tune a Qwen model~\cite{bai2023QwenTechnicalReport}. This fine-tuned model serves as an intelligent evaluation agent, achieving evaluation accuracy comparable to human judgments with minimal error margin.

\subsection{Benchmark Construction}

\noindent \textbf{Multiple-Choice Reading Comprehension} \; 
The first category comprises multiple-choice questions designed to evaluate comprehension of classical Chinese texts. We incorporate reading comprehension tasks for Chinese idioms and passages from WYWEB~\cite{zhouWYWEBNLPEvaluation2023a}. Additionally, we include Traditional Chinese Medicine (TCM) as a significant component of ancient Chinese cultural heritage, primarily documented in classical Chinese. The TCM syndrome differentiation task (TCMSRC) challenges LLMs to select the most appropriate diagnosis from four TCM syndrome types based on patient case descriptions in classical Chinese.

\noindent \textbf{Open-ended Question Answering} \; 
This category evaluates LLMs' deeper understanding and knowledge capabilities by requiring them to generate answers rather than selecting from options, thereby revealing potential hallucinations. The tasks include loan character identification, completion of xiehouyu (traditional Chinese riddles), and TCM-related questions. Leveraging the daizhige database, we constructed tasks requiring LLMs to identify authors and historical dynasties based on book titles, and to categorize books into ten collections based on title comprehension.
Additionally, we developed source attribution tasks for classical Chinese poetry and prose. LLMs must identify the origins of randomly selected poetry verses and famous quotations from ancient texts, including the sources of traditional Chinese idioms. A particularly challenging task is the Inverse Poetry Translation where LLMs must reconstruct classical poems from their modern Chinese translations. It is a task that humans excel at but often challenges AI models.

\noindent \textbf{Comprehension-Based Generation} \; 
This category features generation tasks where output quality is evaluated using BLEU scores to measure similarity with reference answers. Tasks include:
1) Translation of classical poetry and prose into modern Chinese. 2) Comprehensive poetry appreciation. 3) Idiom interpretation. 4) Detailed interpretation of traditional medical prescriptions. 

\noindent \textbf{Pattern-constrained Generation} \;
The final category assesses LLMs' ability to generate text following classical Chinese literary forms and regulations, which have influenced various East Asian literary traditions, including Japanese Haiku. We implement two specific tasks:
\begin{itemize}
    \item  Couplet Generation: Using a comprehensive couplet dataset, LLMs must generate the second sentence given the first sentence, with evaluation based on matching character count.
    
    \item Ci Poetry Generation: Based on an extensive collection of cipai (词牌, formal patterns) and their corresponding rules, LLMs must compose ci poetry according to specified themes and patterns, with scores determined by adherence to formal requirements.
\end{itemize}

\subsection{Evaluation Metrics}

We employ different evaluation strategies for various task types in Fùxì, considering both the nature of the tasks and the practical challenges in assessment.

\noindent \textbf{Multiple-Choice Reading Comprehension.}  \;
For multiple-choice reading comprehension tasks, we use accuracy as the evaluation metric. The answer is extracted from LLM-generated responses using regular expressions. A response is considered incorrect if either the answer extraction fails or the extracted answer differs from the ground truth.

\noindent \textbf{Open-ended Question Answering.} \;
While accuracy remains the primary metric for open-ended QA tasks, the diverse nature of LLM-generated responses poses challenges for automated evaluation. Manual assessment, though accurate, is resource-intensive and difficult to scale. To address this, we develop an LLM-based evaluator following the principles outlined in the LangChain evaluation framework. We first create a training dataset of 2,000 human-annotated examples, where each example consists of evaluation criteria, reference answers, and predicted answers, along with binary correctness labels. This dataset is then used to fine-tune Qwen2-7B-Instruct, resulting in our LLM Evaluator.
To validate the reliability of our LLM Evaluator, we conduct a correlation study between human and automated assessments. We randomly sample 1,000 examples from the evaluated results and calculate the correlation coefficient between human judgments and the LLM Evaluator's decisions. The strong correlation supports the validity of our automated evaluation approach. Details are described in Appendix~\ref{app:llm_evaluator_details}. 

\noindent \textbf{Text Generation.}  \;
For generation tasks, we employ two distinct evaluation strategies based on the task characteristics:
1) For translation-oriented tasks where the target is modern Chinese, we utilize the BLEU metric~\cite{papineni2002bleu}, which effectively captures the semantic and structural similarities between generated and reference translations.
2) For pattern-constrained  generation tasks such as classical poetry and couplets, we focus on evaluating adherence to formal rules (格律). Our evaluation metrics specifically check for correct tonal patterns, rhythmic structures, and parallel construction requirements. While artistic quality is an important aspect of classical Chinese poetry, we prioritize format correctness as the fundamental capability, leaving aesthetic evaluation for future work. This approach allows us to establish a clear baseline for assessing models' ability to generate formally correct classical Chinese text.

\begin{table}[]
    \resizebox{\hsize}{!}{%
    \begin{tabular}{@{}cccc@{}}
    \toprule
    \textbf{Task} & \textbf{\# Questions} & \textbf{\# Avg. QL} & \textbf{ \# Avg. AL} \\ \midrule
    ACRC & 300 & 814.28 & 1.00 \\
    IRC & 1000 & 159.54 & 1.00 \\
    TCMSRC & 1098 & 543.93 & 1.00 \\
    TCMQA & 740 & 385.06 & 47.28 \\
    LCQA & 500 & 37.95 & 16.88 \\
    ASQA & 553 & 19.89 & 5.26 \\
    BA & 384 & 17.37 & 2.64 \\
    BD & 372 & 22.26 & 1.12 \\
    BCC & 532 & 52.65 & 2.00 \\
    PG & 300 & 27.48 & 89.23 \\
    PLST & 400 & 25.22 & 11.46 \\
    FQST & 200 & 29.41 & 10.06 \\
    IST & 962 & 14.17 & 40.61 \\
    IPT & 300 & 252.08 & 152.99 \\
    PT & 300 & 145.79 & 213.91 \\
    ACT & 1000 & 36.33 & 35.97 \\
    PA & 109 & 112.70 & 1312.32 \\
    TCMNER & 300 & 485.71 & 109.46 \\
    IE & 1236 & 14.10 & 20.43 \\
    PE & 404 & 19.74 & 70.11 \\
    CG & 500 & 27.91 & 13.91 \\
    CiG & 300 & 49.81 & 79.35 \\ \bottomrule
    \end{tabular}%
    }
    \caption{Statistics of \textbf{Q}uestion \textbf{L}ength and \textbf{A}nswer \textbf{L}ength for all tasks in Fuxi Benchmark.}
    \label{tab:test_stat}
\end{table}

\begin{table*}[]
    \resizebox{\textwidth}{!}{%
    \begin{tabular}{@{}lccc|cccccccccc|cccccc@{}}
        \toprule
        Model                & ACRC  & IRC   & TCMSRC & ASQA  & LCQA  & BA    & BD    & BCC   & PG    & PLST  & FQST  & IST   & IPT   & PT    & ACT   & PA   & IE    & TCMQA & PE     \\ \midrule
        gpt-4o-mini          & 50.67 & 77.60 & 79.23  & 6.69  & 10.00 & 1.04  & 16.67 & 78.20 & 2.00  & 3.75  & 16.50 & 1.25  & 3.67  & 5.61  & 19.77 & 5.87 & 1.78  & 31.93 & 55.32  \\
        gpt-4o               & 66.33 & 86.00 & 84.79  & 28.57 & 26.80 & 10.16 & 38.98 & 91.73 & 41.33 & 24.00 & 46.00 & 12.89 & 36.33 & 1.80  & 19.86 & 4.88 & 3.02  & 38.12 & 66.07  \\
        qwen-max             & 86.00 & 85.90 & 86.97  & 52.26 & 42.40 & 20.31 & 40.32 & 96.61 & 45.00 & 58.00 & 75.00 & 41.27 & 73.33 & 1.64  & 22.85 & 6.43 & 1.85  & 22.81 & 53.43  \\
        glm-4-plus           & 82.33 & 87.50 & 83.52  & 51.36 & 37.20 & 29.69 & 52.96 & 97.37 & 76.67 & 52.25 & 82.00 & 49.17 & 66.00 & 3.88  & 17.24 & 8.53 & 2.03  & 23.41 & 38.82  \\ \midrule
        internlm25-7b       & 48.67 & 77.20 & 67.85  & 3.62  & 4.20  & 0.00  & 1.88  & 68.61 & 3.00  & 15.75 & 31.50 & 8.52  & 25.33 & 2.07  & 14.08 & 7.49 & 85.53 & 4.32  & 40.38  \\
        internlm25-20b      & 59.33 & 80.70 & 73.95  & 26.40 & 31.80 & 3.39  & 21.51 & 93.23 & 40.67 & 33.25 & 70.00 & 63.20 & 36.00 & 1.81  & 12.51 & 7.53 & 56.86 & 3.78  & 39.19  \\
        glm4-9b             & 63.67 & 77.80 & 76.05  & 2.17  & 5.60  & 1.30  & 14.78 & 73.12 & 2.67  & 8.00  & 28.50 & 3.64  & 25.00 & 9.38  & 18.06 & 7.01 & 1.92  & 26.36 & 56.79  \\
        llama31-8b          & 30.00 & 61.70 & 56.19  & 0.36  & 2.20  & 0.52  & 9.14  & 59.21 & 0.00  & 0.25  & 3.00  & 0.42  & 1.00  & 4.12  & 9.88  & 2.53 & 2.06  & 39.79 & 1.09   \\
        llama31-8b-chinese & 37.00 & 67.50 & 56.01  & 0.00  & 0.60  & 0.00  & 2.69  & 72.37 & 0.33  & 0.25  & 2.50  & 0.10  & 1.00  & 7.32  & 13.72 & 6.21 & 1.07  & 25.29 & 55.33  \\
        qwen2-0.5b          & 34.33 & 46.20 & 28.05  & 0.18  & 0.80  & 0.26  & 3.76  & 12.78 & 1.00  & 1.25  & 2.50  & 0.83  & 13.67 & 4.51  & 14.51 & 2.37 & 2.71  & 31.14 & 1.29   \\
        Xunzi-qwen2-7b     & 45.00 & 62.30 & 54.37  & 12.48 & 2.20  & 0.00  & 5.65  & 19.55 & 28.00 & 5.75  & 26.00 & 23.91 & 11.67 & 1.06  & 1.15  & 3.93 & 20.38 & 2.55  & 22.10  \\
        qwen2-7b            & 77.00 & 79.80 & 82.88  & 5.24  & 3.40  & 1.04  & 8.87  & 74.44 & 0.00  & 13.00 & 22.00 & 2.18  & 21.67 & 1.63  & 15.97 & 6.75 & 1.36  & 27.46 & 47.84  \\
        qwen2.5-1.5b        & 46.67 & 71.70 & 71.13  & 2.17  & 5.40  & 2.08  & 17.20 & 59.77 & 24.67 & 11.00 & 14.50 & 8.21  & 30.33 & 6.91  & 17.47 & 2.66 & 2.61  & 43.64 & 1.10   \\
        qwen2.5-3b          & 57.67 & 77.60 & 76.23  & 18.44 & 12.80 & 3.39  & 18.01 & 83.83 & 28.00 & 18.75 & 41.00 & 8.94  & 41.67 & 7.78  & 20.47 & 6.05 & 1.45  & 27.84 & 61.68  \\
        qwen2.5-7b          & 71.33 & 83.20 & 83.33  & 22.78 & 18.60 & 3.39  & 20.97 & 92.67 & 29.67 & 28.00 & 44.50 & 18.81 & 41.67 & 7.64  & 19.15 & 6.72 & 1.74  & 23.66 & 55.40  \\
        qwen2.5-14b         & 78.67 & 82.50 & 83.15  & 34.72 & 29.60 & 5.47  & 38.44 & 92.67 & 62.67 & 38.00 & 54.00 & 30.25 & 57.33 & 9.28  & 18.21 & 6.67 & 2.35  & 24.95 & 62.65  \\
        qwen2.5-72b         & 83.00 & 85.80 & 86.89  & 35.44 & 31.60 & 14.32 & 43.28 & 93.80 & 48.00 & 46.00 & 61.00 & 42.72 & 77.67 & 3.46  & 24.99 & 7.31 & 2.46  & 22.36 & 41.40  \\ \bottomrule
        \end{tabular}%
    }
    \caption{Evaluation results of selected models on Fuxi benchmark under zero-shot settings. }
    \label{tab:fuxi_zero_shot_main}

\end{table*}

\begin{table}[!t]
    \centering
    \resizebox{\columnwidth}{!}{%
    \begin{tabular}{@{}lcccc@{}}
        \toprule
        \multirow{2}{*}{Model} & \multicolumn{2}{c}{Zero-shot Setting} & \multicolumn{2}{c}{ICL Setting (5-shot)} \\ \cmidrule(l){2-5} 
                               & CG                 & CiG              & CG                  & CiG                \\ \midrule
        gpt-4o-mini            & 79.00              & 2.33             & 71.80               & 10.00               \\
        gpt-4o                 & 89.60              & 55.33             & 83.40               & 53.33               \\
        qwen-max               & 96.60              & 25.67             & 93.80               & 39.67              \\
        glm-4-plus             & 19.20              & 41.67             & 44.00               & 54.33              \\ \midrule
        internlm25-7b         & 32.40              & 17.67             & 59.80               & 18.67              \\
        internlm25-20b        & 63.40              & 3.67             & 44.40               & 6.67               \\
        glm4-9b               & 69.20              & 3.67             & 59.80               & 7.67               \\
        llama31-8b            & 8.60               & 0.33             & 0.20                & 0.00               \\
        llama31-8b-chinese   & 44.80              & 0.67            & 8.40                & 1.00               \\
        qwen2-0.5b            & 32.60              & 0.00             & 8.00                & 0.33               \\
        Xunzi-qwen2-7b       & 0.00               & 0.00             & 0.00                & 0.00               \\
        qwen2-7b              & 62.6               & 7.67              & 36.20               & 10.67               \\
        qwen2.5-1.5b          & 62.60              & 20.00             & 65.80               & 36.67              \\
        qwen2.5-3b            & 56.60              & 31.00             & 12.80               & 13.00               \\
        qwen2.5-7b            & 28.00              & 42.33            & 71.00               & 44.67               \\
        qwen2.5-14b           & 37.60              & 34.67            & 44.00               & 24.33               \\
        qwen2.5-72b           & 65.40              & 60.67            & 85.80               & 53.67              \\ \bottomrule
        \end{tabular}%
    }
    \caption{Evaluation results on Chinese Generation (CG) and Chinese-in-Graph (CiG) tasks under 5-shot settings.}
    \label{tab:exp2_cg_cig}
\end{table}

\section{Experiments}

\subsection{Setup}

The Fùxì Benchmark comprehensively evaluates both knowledge understanding and generation capabilities of large language models (LLMs) in the domain of classical Chinese. For our primary evaluation protocol, we employed a \textbf{zero-shot} setting, where models were provided with task instructions and inputs without exemplars, requiring them to generate answers based solely on their pre-trained knowledge. All task instructions are demonstrated in Appendix~\ref{app:examples}.
Additionally, considering the experimental paradigms used in related studies~\cite{wei2024ACEVALEvaluatingAncient}, we also conducted evaluations using a \textbf{few-shot} in-context learning approach~\cite{brownLanguageModelsAre2020}.

\subsection{Evaluated Models}

We evaluated a variety of state-of-the-art models on the Fùxì Benchmark, including both open-source and commercial closed-source models. For open-source models, the evaluation included the Qwen series~\cite{bai2023QwenTechnicalReport}, GLM-4-9b~\cite{glm2024chatglm}, LLaMA series~\cite{grattafiori2024Llama3Herd}, 
InternLM~\cite{cai2024internlm2}, 
and several fine-tuned variants such as Xunzi~\footnote{\url{https://github.com/Xunzi-LLM-of-Chinese-classics/XunziALLM}} and Chinese-LLaMA~\cite{chinese-llama-alpaca,Llama-Chinese}. 

For closed-source models, we assessed GPT-4o/GPT-4o-min~\cite{openai2023GPT4TechnicalReport}, GLM-4-Plus and Qwen-Max. These models were selected to represent a diverse range of design philosophies and technical capabilities, ensuring a comprehensive evaluation across the Fùxì Benchmark.

\section{Results and Analysis}
\label{sec:results}
\subsection{Overall Performance}
Our comprehensive evaluation across 21 diverse tasks reveals three key insights about LLM capabilities in processing ancient Chinese:

\noindent \textbf{Comprehension-Generation Gap} 
According to Table~\ref{tab:fuxi_zero_shot_main} and Table~\ref{tab:avg_performance}, all models perform significantly better on multiple-choice tasks (ACRC, IRC, TCMSRC) than generation tasks. For example, the average accuracy score on RC tasks is 42.51 higher than that on QA tasks (ASQA to IPT in Table~\ref{tab:fuxi_zero_shot_main}). As for TG tasks (PT to PE in Table~\ref{tab:fuxi_zero_shot_main}) which are evaluated with BLEU score, their lower scores are reasonable given the highly subjective nature of these tasks. However, the wide score range among models also illustrates the substantial performance gap across different architectures.

\noindent \textbf{Performance Scaling} Despite the gap between comprehension and generation, performance on all tasks shows strong correlation with model size, particularly for knowledge-intensive tasks like book-related tasks (BA, BD, BCC) and poetry-related tasks (IPT, PLST). The largest model (72B parameters) outperform its 1.5B counterparts by an average of 22.06\% on RC tasks and 31.85\% on QA tasks, suggesting specialized cultural knowledge benefits significantly from increased parameterization. Performances of the closed-source API models also indicate the scaling effect of model size, with flagship models (GPT-4o, Qwen-Max, GLM-4-Plus) significantly outperforming smaller models (GPT-4o-mini).

\begin{figure}[]
    \centering
        \resizebox{\columnwidth}{!}{
            \includegraphics[scale=0.5]{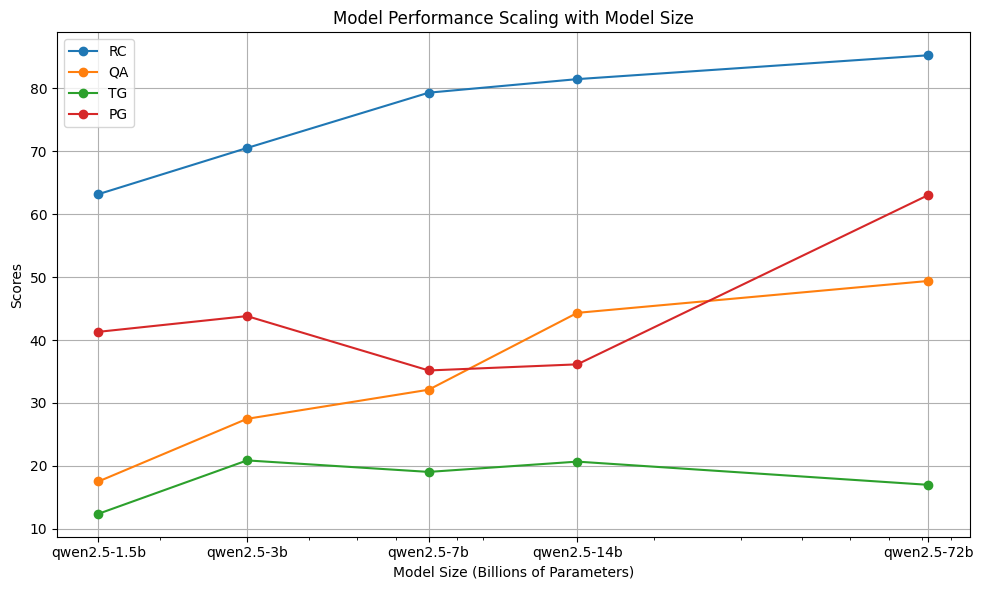}
        }
        \caption{ Model performance scaling with model size on four task categories.}
        \label{fig:performance_scaling}
\end{figure}

\noindent \textbf{Format Constraint Challenge} Models struggle with format-constrained generation, especially LLaMA-31-8b because it is not optimized for Chinese. The extended version (LLaMA-31-8b-Chinese, with optimized Chinese vocabulary and incremental pre-training) shows clear performance improvements across all benchmark tasks. Xunzi-Qwen2-7b, though specialized for ancient Chinese, appears to have lost some instruction-following capability during fine-tuning, leading to an absolute performance drop.
Ci Poetry Generation requires models to utilize their knowledge of cipai (词牌, formal patterns) and generate poetry following specific structures, which proves more difficult than couplet generation. The evaluation results confirm that Couplet Generation is relatively easier, with model average scores 29.46\% higher than on Ci Generation.
GPT-4o and Qwen2.5-72b are the only two models achieving non-trivial performance on Ci Generation under zero-shot settings, indicating their superior adherence to classical Chinese structural constraints. However, most models score below 10\% on this task.
This suggests that adherence to classical Chinese structural constraints presents a unique challenge for current LLMs.

\subsection{Task-Centric Analysis}

\noindent \textbf{Comprehension Tasks}
As shown in Table~\ref{tab:fuxi_zero_shot_main} and Figure~\ref{fig:performance_scaling}, comprehension tasks demonstrate clear scaling advantages, with larger models consistently outperforming smaller variants. The Ancient Chinese RC task (ACRC) shows particular sensitivity to model architecture, with Chinese-originated models (GLM and Qwen series) achieving better performance than LLaMA series, even for API models where Qwen-Max and GLM-4-Plus show a 10\%+ gap compared to GPT-4o.
The commercial models demonstrate superior performance on comprehension tasks, potentially indicating advantages from their larger parameter sizes.

\noindent \textbf{Generation Tasks}

For open-ended generation tasks, larger models clearly perform better. Many of these tasks (involving classical books, poetry, allegorical sayings, loan characters) are knowledge-intensive, particularly source tracing tasks, requiring more parameters to memorize and understand this specialized knowledge. Error analysis reveals that smaller models are more prone to hallucination phenomena.

For text generation tasks, there is no clear scaling pattern in Figure~\ref{fig:performance_scaling}. This is because these tasks can be divided into translation tasks (PT, ACT) and appreciation/explanation tasks (PA, IE, PE, TCMQA), all using BLEU as the evaluation metric with reference answers that may not be optimal. Nevertheless, models with more parameters still perform better on ancient Chinese short text translation. Notably, InternLM and Xunzi-Qwen2-7b perform exceptionally well on the Idiom Explanation (IE) task, possibly because their pre-training data included more relevant texts.

For pattern-constrained generation tasks, the results are evaluated using rule-based metrics since LLM evaluators cannot precisely assess format requirements, resulting in generally lower scores. Xunzi-Qwen2-7b scored 0 because its outputs contained many irrelevant descriptions or failed to stop generating. Qwen2-0.5 similarly scored low because it couldn't maintain the required pattern rules. LLaMA-31-8b exhibited language mixing issues, and while LLaMA-31-8b-Chinese greatly improved this aspect, it still struggled with pattern adherence.

\begin{table}[]
    \centering
    \resizebox{\columnwidth}{!}{%
    \begin{tabular}{@{}lcccc@{}}
        \toprule
        \textbf{Model} &  \textbf{RC} &  \textbf{QA} & \textbf{TG} & \textbf{PG} \\ \midrule
        gpt-4o-mini    & 69.17 & 13.98 & 20.05 & 40.67 \\
        gpt-4o         & 79.04 & 35.68 & 22.29 & 72.47 \\
        qwen-max       & 86.29 & 54.45 & 18.17 & 61.14 \\
        glm-4-plus     & 84.45 & 59.47 & 15.65 & 30.44 \\
        internlm25-7b  & 64.57 & 16.24 & 25.65 & 25.04 \\
        internlm25-20b & 71.33 & 41.95 & 20.28 & 33.54 \\
        glm4-9b        & 72.51 & 16.48 & 19.92 & 36.44 \\
        llama31-8b     & 49.30 & 7.61  & 9.91  & 4.47  \\
        llama31-8b-chinese & 53.50 & 7.98  & 18.16 & 22.74 \\
        qwen2-0.5b     & 36.19 & 3.70  & 9.42  & 16.30 \\
        Xunzi-qwen2-7b & 53.89 & 13.52 & 11.50 & 0.00  \\
        qwen2-7b       & 79.89 & 15.18 & 16.84 & 35.14 \\
        qwen2.5-1.5b   & 63.17 & 17.53 & 12.40 & 41.30 \\
        qwen2.5-3b     & 70.50 & 27.48 & 20.88 & 43.80 \\
        qwen2.5-7b     & 79.29 & 32.11 & 19.05 & 35.17 \\
        qwen2.5-14b    & 81.44 & 44.32 & 20.69 & 36.14 \\
        qwen2.5-72b    & 85.23 & 49.38 & 17.00 & 63.04 \\ \bottomrule
        \end{tabular}
        
    }
    \caption{Average performance of models across task categories:  Multiple-choice Reading Comprehension (RC), Open-ended Question Answering (QA), Text Generation(TG) and Pattern-constrained Generation(PG).}
    \label{tab:avg_performance}
\end{table}

\subsection{LLM Evaluator Validation}
\label{app:llm_eval}

\begin{figure}[ht]
    \centering
    \begin{tikzpicture}
    \begin{axis}[
        width=\columnwidth,
        height=0.6\columnwidth,
        ybar,
        bar width=0.15cm,
        ylabel={Score},
        xlabel={},
        symbolic x coords={ASQA, BA, BCC, BD, IST, LCQA, PG, IPT, PLST, FQST},
        xtick=data,
        xticklabel style={rotate=45, anchor=east, font=\footnotesize},
        legend style={at={(0.5,1.25)}, anchor=north, legend columns=-1},
        ymin=0, ymax=1,
        grid=major,
        ]
    \addplot coordinates {
        (ASQA, 0.870)
        (BA, 0.953)
        (BCC, 0.899)
        (BD, 0.976)
        (IST, 0.930)
        (LCQA, 0.963)
        (PG, 0.448)
        (IPT, 0.955)
        (PLST, 0.966)
        (FQST, 0.867)
    };
    \addplot coordinates {
        (ASQA, 0.567)
        (BA, 0.775)
        (BCC, 0.646)
        (BD, 0.950)
        (IST, 0.581)
        (LCQA, 0.903)
        (PG, 0.000)
        (IPT, 0.898)
        (PLST, 0.905)
        (FQST, 0.682)
    };
    \legend{Accuracy,Cohen's $\kappa$}
    \end{axis}
    \end{tikzpicture}
    \caption{Comparison of Accuracy and Cohen's Kappa across different labels. Overall performance: Accuracy = 0.898, Cohen's Kappa = 0.764.}
    \label{fig:acc_kappa}
    \end{figure}
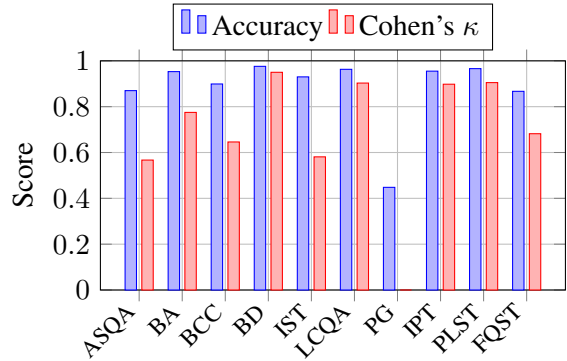

\begin{table}[ht]
    \centering

    \begin{tabular}{lcc}
    \hline
    \textbf{Metric} & \textbf{w/ PG} & \textbf{w/o PG} \\
    \hline
    Accuracy & 0.898 & 0.930 \\
    Cohen's $\kappa$ & 0.764 & 0.830 \\
    Precision & 0.833 & 0.815 \\
    Recall & 0.843 & 0.951 \\
    F1 Score & 0.838 & 0.878 \\
    Pearson Correlation & 0.764 & 0.835 \\
    \hline
    \end{tabular}
    \caption{Overall Performance Metrics. w/o PG: without Poetry Generation (PG) task.}
    \label{tab:performance_metrics}
    \end{table}

To ensure reliable evaluation of open-ended generation tasks, we validated our LLM-based evaluator through a comprehensive human evaluation study. We randomly sampled 1,000 responses from GPT-4o's outputs across different task categories, with sampling proportions matching the distribution of tasks in the benchmark. Three expert annotators independently judged each response, reaching consensus on each sample to establish gold standard answers. As shown in Figure 5, our evaluator demonstrates strong alignment with human judgments, achieving overall accuracy of 89.8\% and substantial agreement (Cohen's $\kappa$ = 0.764).
The evaluator performs exceptionally well on factual tasks like Book Dynasty identification (BD, $\kappa$ = 0.950) and Poetry Line Source Tracing (PLST, $\kappa$ = 0.905). It shows slightly lower agreement on subjective tasks like Allegorical Saying QA (ASQA, $\kappa$ = 0.567), mirroring patterns in human inter-annotator agreement. As shown in Table 5, excluding the Poetry Generation (PG) task improves overall agreement to $\kappa$ = 0.830. This exclusion was necessary because our aligned LLM evaluator incorrectly penalized valid responses providing three candidate poetic forms (体裁), confirming that our automated evaluation approach provides reliable assessment for most tasks.
The strong correlation between human and automated assessments (Pearson's r = 0.835 excluding PG) validates our evaluation methodology, enabling consistent comparison across models and tasks while minimizing resource-intensive manual evaluation.

\section{Conclusion}

This paper introduces Fùxì, a comprehensive benchmark for evaluating large language models' capabilities in ancient Chinese text understanding and generation. Through extensive evaluation of current state-of-the-art models across 21 diverse tasks, our study reveals several important findings. First, while modern LLMs demonstrate promising performance in basic comprehension tasks, they still face significant challenges in tasks requiring deep cultural understanding and specialized knowledge. Second, the generation of authentic ancient Chinese text remains a particularly challenging area, with models struggling to maintain consistent classical style and adhere to traditional formats. Third, model size and architecture choices significantly impact performance, with larger models and those specifically trained on Chinese corpora showing advantages in certain tasks.
Beyond these empirical findings, Fùxì contributes to the field by providing: (1) a standardized evaluation framework that encompasses both understanding and generation capabilities, (2) novel evaluation metrics addressing classical Chinese's unique structural requirements, validated through human-LLM evaluator alignment, and (3) baseline results that can guide future research and development in this area. The insights gained from this benchmark can inform both the development of more effective models for processing ancient Chinese texts and the broader field of cultural heritage preservation through AI.

\section{Limitations}


\noindent \textbf{Evaluation Metrics} 

Although we introduce novel metrics for generation tasks, the automatic evaluation of generated ancient Chinese text remains challenging. Current metrics may not fully capture the nuances of classical Chinese style and literary conventions. Future work could explore more sophisticated evaluation approaches that better align with human expert judgments.

\noindent \textbf{Task Coverage} 

While our benchmark covers a broad range of tasks, certain aspects of ancient Chinese language processing remain underexplored. For instance, the benchmark could be expanded to include more specialized genres of classical literature or more diverse historical periods. Additionally, the current focus on written text does not address the phonological aspects of ancient Chinese.

These limitations point to several promising directions for future research, including the development of more sophisticated evaluation metrics, expansion of task coverage, and investigation of methods to address historical data biases. We hope that acknowledging these limitations will stimulate further research and improvements in ancient Chinese language processing.


\bibliography{custom}

\appendix

\clearpage

\section{ICL Evaluation}
\label{sec:appendix}

The results of 5-shot in-context learning setting are shown in Table~\ref{tab:fuxi_icl_main}. 
Under in-context learning setting, each sample is prompted with 5 examples from the development set. The quantitative statistics of development set are shown in Table~\ref{tab:fuxi_dev_stat}.

\begin{table*}[thbp]
    \resizebox{\textwidth}{!}{%
    \begin{tabular}{lccc|cccccccccc|cccccc}
        \toprule
        Model                & ACRC  & IRC   & TCMSRC & ASQA  & LCQA  & BA    & BD    & BCC   & PG    & PLST  & FQST  & IST   & IPT   & PT    & ACT   & PA   & IE    & TCMQA & PE    \\ \midrule
        glm-4-plus           & 78.67 & 88.40 & 88.80  & 42.68 & 46.40 & 42.97 & 67.20 & 79.89 & 63.00 & 45.75 & 77.00 & 51.66 & 34.67 & 16.10 & 26.57 & 9.24 & 8.56  & 48.85 & 3.58  \\ \midrule
        internlm25\_7b       & 51.00 & 68.20 & 51.64  & 6.33  & 16.60 & 0.78  & 22.04 & 48.50 & 10.00 & 3.50  & 10.50 & 2.49  & 23.00 & 1.92  & 7.06  & 6.94 & 68.53 & 9.24  & 1.73  \\
        internlm25\_20b      & 43.67 & 64.90 & 73.22  & 8.68  & 7.40  & 4.69  & 6.45  & 64.10 & 13.00 & 4.75  & 37.50 & 10.50 & 32.33 & 3.48  & 9.88  & 5.58 & 81.63 & 6.11  & 2.33  \\
        glm4\_9b             & 40.33 & 75.10 & 64.75  & 6.51  & 18.80 & 11.46 & 40.32 & 75.38 & 26.00 & 4.25  & 35.00 & 11.95 & 32.00 & 5.11  & 19.72 & 6.83 & 4.35  & 39.13 & 7.12  \\
        llama31\_8b          & 27.00 & 25.80 & 26.59  & 3.98  & 1.60  & 0.78  & 6.72  & 56.20 & 4.00  & 0.00  & 0.50  & 3.12  & 2.67  & 55.14 & 3.39  & 5.02 & 76.29 & 40.62 & 3.04  \\
        llama31\_8b\_chinese & 28.33 & 15.30 & 37.89  & 8.32  & 2.20  & 2.34  & 20.70 & 64.29 & 23.00 & 0.00  & 4.00  & 6.24  & 4.00  & 28.09 & 12.50 & 4.59 & 75.93 & 32.65 & 3.21  \\
        qwen2\_0.5b          & 24.67 & 27.10 & 25.14  & 1.63  & 1.80  & 1.56  & 6.99  & 8.65  & 1.67  & 0.75  & 5.00  & 2.39  & 3.67  & 1.33  & 4.16  & 2.70 & 1.77  & 31.50 & 2.37  \\
        qwen2\_1.5b          & 39.33 & 64.00 & 45.63  & 4.70  & 1.80  & 2.08  & 11.83 & 35.53 & 4.00  & 3.50  & 10.50 & 3.43  & 9.67  & 4.35  & 11.73 & 4.58 & 4.20  & 34.27 & 5.30  \\
        qwen2\_7b            & 63.67 & 83.40 & 80.05  & 15.73 & 16.60 & 6.51  & 29.84 & 75.00 & 21.33 & 17.50 & 29.50 & 12.68 & 28.33 & 4.41  & 20.38 & 6.65 & 3.71  & 43.07 & 4.89  \\
        Xunzi\_qwen2\_7b     & 34.67 & 62.40 & 50.09  & 1.81  & 0.40  & 1.56  & 5.65  & 16.73 & 13.33 & 3.50  & 12.50 & 9.77  & 6.00  & 1.28  & 78.06 & 5.50 & 15.23 & 2.66  & 48.28 \\
        qwen2.5\_1.5b        & 41.67 & 37.80 & 46.81  & 3.98  & 3.80  & 4.69  & 33.87 & 49.44 & 9.67  & 11.75 & 19.50 & 5.09  & 14.67 & 7.50  & 19.23 & 5.06 & 5.85  & 46.39 & 6.66  \\
        qwen2.5\_3b          & 58.00 & 73.00 & 73.22  & 22.97 & 5.00  & 2.60  & 29.03 & 61.09 & 26.67 & 20.50 & 28.50 & 17.26 & 24.00 & 8.04  & 22.00 & 6.93 & 2.71  & 39.08 & 6.73  \\
        qwen2.5\_7b          & 73.00 & 82.30 & 83.97  & 30.56 & 18.40 & 11.46 & 33.60 & 82.52 & 35.00 & 30.00 & 34.00 & 15.28 & 41.00 & 7.16  & 24.27 & 7.76 & 2.70  & 42.98 & 4.10  \\
        qwen2.5\_14b         & 72.33 & 85.30 & 85.15  & 38.16 & 21.20 & 12.24 & 35.22 & 92.67 & 54.00 & 28.00 & 60.00 & 35.65 & 28.33 & 10.65 & 21.72 & 7.25 & 3.62  & 33.29 & 2.98  \\
        qwen2.5\_72b         & 83.33 & 85.60 & 83.97  & 49.55 & 35.20 & 22.14 & 47.85 & 89.47 & 45.67 & 50.00 & 58.00 & 36.17 & 66.67 & 12.35 & 28.14 & 8.53 & 5.98  & 30.16 & 7.81  \\
        yi34bchat            & 61.33 & 81.90 & 75.14  & 18.26 & 8.60  & 5.99  & 40.32 & 69.92 & 20.33 & 13.75 & 30.00 & 28.48 & 49.67 & 11.61 & 22.09 & 6.31 & 5.42  & 23.66 & 8.11  \\ \bottomrule
        \end{tabular}%
    }
    \caption{Evaluation results of selected models on Fuxi benchmark under 5-shot in-context learning settings.}
    \label{tab:fuxi_icl_main}
\end{table*}

\begin{table}[!thbp]
    \centering
    \resizebox{\columnwidth}{!}{%
    \begin{tabular}{@{}lcccc@{}}
        \toprule
        \textbf{Model} &  \textbf{RC} &  \textbf{QA} & \textbf{TG} & \textbf{PG} \\ \midrule
        gpt-4o-mini    & - & - & - & 40.90 \\
        gpt-4o         & - & - & - & 68.37 \\
        qwen-max       & - & - & - & 66.74 \\
        glm-4-plus     & 85.29 & 55.12 & 18.82 & 49.17 \\
        internlm25-7b  & 56.95 & 14.37 & 15.90 & 39.24 \\
        internlm25-20b & 60.60 & 18.94 & 18.17 & 25.54 \\
        glm4-9b        & 60.06 & 26.17 & 13.71 & 33.74 \\
        llama31-8b     & 26.46 & 7.96  & 30.58 & 0.10  \\
        llama31-8b-chinese & 27.17 & 13.51 & 26.16 & 4.70  \\
        qwen2-0.5b     & 25.64 & 3.41  & 7.31  & 4.17  \\
        Xunzi-qwen2-7b & 49.05 & 7.13  & 25.17 & 0.00  \\
        qwen2-7b       & 75.71 & 25.30 & 13.85 & 23.44 \\
        qwen2.5-1.5b   & 42.09 & 15.65 & 15.12 & 51.24 \\
        qwen2.5-3b     & 68.07 & 23.76 & 14.25 & 12.90 \\
        qwen2.5-7b     & 79.76 & 33.18 & 14.83 & 57.84 \\
        qwen2.5-14b    & 80.93 & 40.55 & 13.25 & 34.17 \\
        qwen2.5-72b    & 84.30 & 50.07 & 15.50 & 69.74 \\ \bottomrule
        \end{tabular}
        
    }
    \caption{Average performance of models across task categories under 5-shot in-context learning settings: Multiple-choice Reading Comprehension (RC), Open-ended Question Answering (QA), Text Generation (TG) and Pattern-constrained Generation (PG).}
    \label{tab:avg_icl_performance}
\end{table}

\begin{table}[]
    \resizebox{\hsize}{!}{%
    \begin{tabular}{@{}cccc@{}}
    \toprule
     \textbf{Task} & \textbf{\# Questions} & \textbf{\# Avg. AL} & \textbf{ \# Avg. AL} \\ \midrule
    ACRC   &     5  &        857. 0 &                 1. 0 \\
    IRC    &     5  &        152. 2 &                 1. 0 \\
    TCMSRC &     5  &        677. 6 &                 1. 0 \\
    TCMQA  &     5  &        325. 8 &                40. 6 \\
    LCQA   &     5  &         44. 6 &                16. 0 \\
    ASQA   &     5  &         20. 2 &                 5. 0 \\
    BA     &     5  &         23. 8 &                 2. 4 \\
    BD     &     5  &         23. 8 &                 1. 2 \\
    BCC    &     5  &         52. 8 &                 2. 0 \\
    PG     &     5  &         27. 4 &                62. 6 \\
    PLST   &     5  &         25. 4 &                11. 0 \\
    FQST   &     5  &         32. 0 &                 9. 6 \\
    IST    &     5  &         14. 0 &                47. 4 \\
    IPT    &     5  &       2420. 8 &              1464. 0 \\
    PT     &     5  &        140. 0 &               206. 6 \\
    ACT    &     5  &         36. 4 &                38. 6 \\
    PA     &     5  &         63. 6 &              1100. 0 \\
    TCMNER &     5  &        403. 2 &               117. 2 \\
    IE     &     5  &         14. 0 &                16. 2 \\
    PE     &     5  &         19. 4 &                63. 8 \\
    CG     &     5  &         27. 4 &                13. 4 \\
    CiG    &     5  &         46. 6 &                60. 8 \\ \bottomrule
    \end{tabular}%
    }
    \caption{Quantitative statistics of \textbf{Q}uestion \textbf{L}ength and \textbf{A}nswer \textbf{L}ength in development set. }
    \label{tab:fuxi_dev_stat}
\end{table}

\subsection{In-context Learning Results}

Compared to zero-shot performance, we observe several interesting patterns:

First, in-context learning provides asymmetric benefits across model scales. Smaller models (<7B parameters) show substantial improvements with 5-shot examples, with an average gain of 27.4 percentage points on tasks like Couplet Generation (CG). For instance, internlm25\_7b improves from 32.4\% to 59.8\% on pattern-constrained generation tasks when provided with examples.

Second, larger models (>20B parameters) benefit less from in-context examples, with an average improvement of only 5.9 percentage points across generation tasks. This suggests that increased model size reduces context-dependence, as larger models have already internalized more knowledge during pre-training.

Third, the performance gap between comprehension and generation tasks persists even with in-context examples. While multiple-choice tasks like ACRC see modest improvements (average +4.2 points), challenging generation tasks like Ci Poetry Generation (CiG) remain difficult even with examples, with only qwen2.5-72b achieving non-trivial performance (24.0\%).

Finally, we observe that domain-specific tasks benefit more from in-context examples than general tasks. Traditional Chinese Medicine tasks (TCMSRC, TCMQA) show an average improvement of 12.3 percentage points with 5-shot examples, suggesting that specialized knowledge can be effectively activated through demonstration.

\section{LLM Evaluator Details}
\label{app:llm_evaluator_details}

To evaluate the open-ended question answering tasks automatically, LLMs are good choices as we have witnessed the their impressive performances on some of our tasks. However, directly using LLMs to evaluate open-ended question answering tasks is not a good idea as the LLMs are not reliable. 
To address this issue, we first design a LLM-based evaluator following the principles outlined in the LangChain~\footnote{https://github.com/langchain-ai/langchain} evaluation framework. Unfortunatly, even under the langhcain framework, the LLM evaluator is still not reliable on judge's role.

\begin{figure*}[t]
\begin{tcolorbox}[width=\textwidth, colframe=blue!50!black, colback=blue!5!white, sharp corners]

\vspace{0.5em}

\small
\texttt{作为一名阅卷老师，请根据以下评分标准对以下回答进行评价，评价结果的形式是：如果回答正确，输出"Y" ，如果回答错误输出"N"。绝对不要输出额外的内容。}

\texttt{评分标准:}

\begin{itemize}
    \item \texttt{包含要点: 理解问题和标准回答，之后从标准答案中提取出一个答案要点，预测答案必须包含这个要点。}
    \item \texttt{与标准答案一致: 如果预测答案是一段文字，那么预测答案必须与预标准答案传达的信息一致。}
    \item \texttt{符合问题和任务要求: 如果问题要求创作诗词, 那么回答中的诗词创作必须符合要求的主题和格式。基于对作品的理解进行评判。}
    \item \texttt{符合事实: 如果问题要求给出古诗词原文, 那么回答给出的诗句必须是历史上存在的，而且必须和标准答案一致。}
    \item \texttt{预测答案简洁明了: 预测答案需要简洁明了，如果出现重复内容或非中文文本则直接判断为错误。}
\end{itemize}

\texttt{<start>}

\texttt{***}

\texttt{[任务输入]: \{input\}}

\texttt{***}

\texttt{[标准答案]: \{output\}}

\texttt{***}

\texttt{[预测答案]: \{prediction\}}

\texttt{***}

\texttt{<end>}

\texttt{[评价结果]:\textbackslash n Y/N \textbackslash n 判断理由: \{reason\}}

\end{tcolorbox}
\caption{Prompt for the LLM Evaluator. }
\label{fig:llm_evaluator_prompt}
\end{figure*}

To this end, we further design a training dataset of 2,000 human-annotated examples, where each example consists of evaluation criteria, task input, reference answers, and predicted answers, along with binary correctness labels. Furthermore, the model is required to output its own reasoning for the judgments. The prompt template is demonstrated in Figure~\ref{fig:llm_evaluator_prompt}.
We use the dataset to fine-tune Qwen2-7B-Instruct, resulting in our LLM Evaluator.

\section{Case Study and Error Analysis}

\begin{table*}[htbp]
\begin{tabular}{|p{\textwidth}|}
\hline
\textbf{Task: Inverse Poetry Translation} \\
\hline
\textbf{Instruction:} 
根据以下翻译给出对应的古诗词原文。\\
\textcolor{translationcolor}{Give the corresponding original text of ancient poetry based on the following translation.} \\
\hline
\textbf{Input:} \\
听一支新曲喝一杯美酒，还是去年的天气旧日的亭台，西落的夕阳何时再回来？\\
那花儿落去我也无可奈何，那归来的燕子似曾相识，在小园的花径上独自徘徊。\\
\textcolor{translationcolor}{Listening to a new song and drinking a glass of fine wine, it is still the same weather as last year and the same pavilion as before. When will the setting sun come back?} \\
\textcolor{translationcolor}{I can do nothing about the falling flowers, and the returning swallows seem familiar, wandering alone on the flower path in the small garden.} \\ \hline
\textbf{Output:} \\
题目: 浣溪沙·一曲新词酒一杯 \textcolor{translationcolor}{Title: Huanxisha·A new song and a glass of wine} \\
作者: 晏殊   \textcolor{translationcolor}{Author: Yan Shu}\\
一曲新词酒一杯，去年天气旧亭台。夕阳西下几时回？\\
无可奈何花落去，似曾相识燕归来。小园香径独徘徊。\\
\textcolor{translationcolor}{A new song and a glass of wine, the weather and pavilions of last year. When will the setting sun return? } 
\textcolor{translationcolor}{The flowers fall helplessly, and the familiar swallows return. I wander alone in the fragrant path of the small garden.}\\ \hline
\textbf{GPT-4o} \textcolor{correctcolor}{Correct} \\ 
这段翻译对应的古诗词是宋代晏殊的《浣溪沙》："**一曲新词酒一杯，去年天气旧亭台。夕阳西下几时回？无可奈何花落去，似曾相识燕归来。小园香径独徘徊。**" \\ \hline 
\textbf{GPT-4o-mini} \\
这段翻译对应的古诗词原文是《酒泉子·听一曲新词》。这首词是宋代词人温庭筠所作。原文如下:\\
\textcolor{red}{听一曲新词，饮一杯旧酒。} 
去年天气旧亭台，夕阳西下几时回？ \\ 
无可奈何花落去，似曾相识燕归来。 
小园香径独徘徊。\\ \hline 
\textbf{llama31-8b} \\
秋夕憶舊時。\\ \hline 
\textbf{llama31-8b-chinese} \\ 
这段文本并不是直接引用的古典诗词，而是一种现代的表达方式。它表达了一种对往昔美好时光的怀念和对未来无法重复的事物的无奈。这样的表达手法在现代文学中很常见，但不是古典诗词的典型风格。 \\
如果要找到古诗词中的类似意境，可以参考以下几首诗： ...  \\ \hline 
\textbf{glm4-9b} \\ 
独立小桥风细细，旧时王谢堂前燕，飞入寻常百姓家。 \\
此句诗意大致与给出的翻译相符，原文出自唐代诗人刘禹锡的《陋室铭》。原文中的“独立小桥风细细”对应“听一支新曲喝一杯美酒”，“旧时王谢堂前燕”对应“那归来的燕子似曾相识”，“飞入寻常百姓家”则与“那花儿落去我也无可奈何，那归来的燕子似曾相识，在小园的花径上独自徘徊”相呼应。但由于原文中没有直接提到“西落的夕阳何时再回来”，所以这部分内容在原文中并没有直接对应。 \\ \hline 
\textbf{qwen2.5-3b} \\ 
根据您提供的翻译，对应的古诗词原句应该是：
“去年今日此门中，人面桃花相映红。人面不知何处去，桃花依旧笑春风。”
这是一首唐人经典的短诗，作者是崔护，出自《题都城南庄》。您的原句也表达了相似的情感和意境，但这首诗更为人所熟知。 \\ \hline 
\textbf{Xunzi-qwen2-7b} \\ 
 回答:《游园不值》（宋·朱熹） 山光雨欲满谷，绿绕暗溪乔木。 ... 《春宵曲》（宋·苏轼） 春宵一刻值千金，花落人亡两不知。 问君能有几多愁，恰似一江春水向东流。 《如梦令》（宋·李清照）... 回答:以上是我的演讲稿，希望对您有所帮助。祝您学习愉快！请注意，文字中有两 \\ \hline 
\textbf{qwen2.5-72b} \\ 
听新曲，饮美酒，依旧是去年的天气、旧日的亭台，西下的夕阳何时才能回来？
无可奈何花落去，似曾相识燕归来，小园香径独徘徊。 \\ \hline

\end{tabular}
\caption{Error Case for Task: Inverse Poetry Translation}
\label{tab:error_poem_nmt_inv}
\end{table*}

As shown in Table~\ref{tab:error_poem_nmt_inv}, we take some examples of \textit{Inverse Poetry Translation} task from the zero-shot evaluation results. 
This is a super-easy task for human beings, but it is still challenging for LLMs. 
Given the modern Chinese version of the poem, human beings can easily find the corresponding ancient Chinese version.
However, LLMs either fails to generate a poem-like content(llama31-8b, qwen2.5-72b) or generates a wrong poem(llama31-8b-chinese, glm4-9b, qwen2.5-3b). Xunzi-qwen2-7b generates many irrelevant content and show a topic shift in the end. GPT-4o-mini generates most of the content correctly, but it still fails on some tiny mistakes. This phenomenon is also observed across all models.

\section{Task Examples}
\label{app:examples}

The examples of all tasks in Fuxi benchmark are demonstrated below.

\begin{figure*}[t]
\begin{tcolorbox}[
    width=\textwidth, 
    colframe=blue!50!black, 
    colback=blue!5!white, 
    sharp corners, 
    ]
    \textbf{Task: Ancient Chinese RC}\\
    \textbf{Instruction:} \\以下是一道文言文阅读理解的单项选择题, 请直接给出正确答案的选项。\\
    \textcolor{translationcolor}{The following is a multiple-choice question on ancient Chinese reading comprehension. Please provide the correct answer option directly. } \\
    \textbf{Input:}\\
    文章：古之学者必有师。师者, 所以传道受业解惑也。人非生而知之者, 孰能无惑？惑而不从师, 其为惑也, 终不解矣。生乎吾前其闻道也固先乎吾吾从而师之生乎吾后其闻道也亦先乎吾吾从而师之吾师道也夫庸知其年之先后生于吾乎？是故无贵无贱, 无长无少, 道之所存, 师之所存也。 \\
    嗟乎！师道之不传也久矣！欲人之无惑也难矣！古之圣人, 其出人也远矣, 犹且从师而问焉；今之众人, 其下圣人也亦远矣, 而耻学于师。是故圣益圣, 愚益愚。圣人之所以为圣, 愚人之所以为愚, 其皆出于此乎？\\
    爱其子, 择师而教之；于其身也, 则耻师焉, 惑矣。彼童子之师, 授之书而习其句读者, 非吾所谓传其道解其惑者也。句读之不知, 惑之不解, 或师焉, 或不焉, 小学而大遗, 吾未见其明也。巫医乐师百工之人, 不耻相师。士大夫之族, 曰师曰弟子云者, 则群聚而笑之。问之, 则曰："彼与彼年相若也, 道相似也。位卑则足羞, 官盛则近谀。"呜呼！师道之不复, 可知矣。巫医乐师百工之人, 君子不齿, 今其智乃反不能及, 其可怪也欤！\\
    圣人无常师。孔子师郯子、苌弘、师襄、老聃。郯子之徒, 其贤不及孔子。孔子曰：三人行, 则必有我师。是故弟子不必不如师, 师不必贤于弟子, 闻道有先后, 术业有专攻, 如是而已。\\
    李氏子蟠, 年十七, 好古文, 六艺经传皆通习之, 不拘于时, 学于余。余嘉其能行古道, 作《师说》以贻之。\\ \;  \textcolor{translationcolor}{Article: In ancient times those who wanted to learn would seek out a teacher. A teacher's role is to transmit the Way (传道), impart knowledge (受业), and resolve doubts (解惑). Since no one is born omniscient, who can claim to have no doubts? If one has doubts and is not willing to learn from a teacher, his doubts will never be resolved. Anyone who was born before me and learned the Way before me is my teacher. Anyone who was born after me and learned the Way before me is also my teacher. Since what I desire to learn is the Way, why should I care whether he was born before or after me? Therefore, it does not matter whether a person is high or low in position, young or old in age. Where there is the Way, there is my teacher.} \\
    \textcolor{translationcolor}{Alas! The tradition of honoring teachers has long decayed, making it arduous to dispel human confusion. The sages of antiquity, far surpassing ordinary people, still learned from teachers. Modern masses, vastly inferior to sages, shamefully disdain learning from others. Hence, sages grow wiser, fools grow duller. Is this not the root of their divergence? }\\
    \textcolor{translationcolor}{Parents hire tutors for children but shun self-improvement—a paradox. A child's tutor drills texts, not transmitting the Way or resolving profound doubts. Mastering punctuation yet ignoring truth prioritizes trivia over substance. Artisans freely learn, yet elites mock teacher-disciple bonds as "humiliating if inferior, sycophantic if superior." How ironic that despised artisans now surpass elites in wisdom!} \\
    \textcolor{translationcolor}{Sages have no fixed teachers. Confucius learned from Tanzi, Chang Hong, Shi Xiang, and Lao Dan—none surpassing him in virtue. Confucius said: "Among three companions, one must be my teacher." Thus, disciples need not be inferior to teachers, nor teachers superior to disciples. Comprehension of the Way varies in timing; expertise differs in specialization—this is all. 
    Li Pan, seventeen, devoted to classical studies and mastering the Six Arts, learns from me unbound by contemporary prejudices. I commend his adherence to ancient principles and compose this Discourse on Teachers for him.} \\  

\end{tcolorbox}
\caption{Example for task: Ancient Chinese RC (Part 1).}
\end{figure*}

\begin{figure*}[t]\ContinuedFloat
\begin{tcolorbox}[
    width=\textwidth,
    colframe=blue!50!black,
    colback=blue!5!white,
    sharp corners,
]
    题目: 下列对原文有关内容的概括和分析，不正确的一项是:\\A. 第一段所提出的无论贵贱长幼、有道者皆可为师的择师标准，很有积极意义。\\B. 第二段以对比的方法针砭时弊，批判了"士大夫之族"耻于从师的恶劣风气。\\C. 第三段以孔子从师为例，阐明教师的作用和"道之所存，师之所存"的道理。\\D. 第四段作者借赞扬李蟠``不拘于时\textquotedblright ``能行古道\textquotedblright，补充交代本文的写作缘由。\\ \; 
    \textcolor{translationcolor}{Question: Which analysis of the text is incorrect? \\
    A. The first paragraph establishes a progressive standard for selecting teachers---merit lies in the Way, transcending hierarchy---which remains socially impactful.\\
    B. The second paragraph critiques societal decay through contrast, condemning the literati-official class's shame toward learning from teachers.\\
    C. The third paragraph uses Confucius's pedagogical examples to illustrate teachers' roles and the principle \textquotedblleft where the Way resides, there lies the teacher\textquotedblright.\\
    D. The fourth paragraph highlights Li Pan's defiance of trends and revival of ancient practices, contextualizing the essay's purpose.} \\ 
    \textbf{Output:}\\
    C \\
\end{tcolorbox}
\caption{Example for task: Ancient Chinese RC (Part 2).}
\end{figure*}

\begin{figure*}[t]
\begin{tcolorbox}[width=\textwidth, colframe=blue!50!black, colback=blue!5!white, sharp corners]
    \textbf{Task: Idiom RC}\\
    \textbf{Instruction:} \\以下是关于成语的单项选择题，请直接给出正确答案的选项。\\
    \textcolor{translationcolor}{The following is an idiom-based multiple-choice question. Provide the correct answer option directly.} \\
    \textbf{Input:}\\
    题目: \textquotedblleft 祸福无门，唯人所召\textquotedblright。 中的``祸福无门，唯人所召\textquotedblright 的意思是？\\ 
    A. 重遭祸事，是因为自己没有道德而造成的。\\ 
    B. 指灾祸和幸福不是注定的，都是人们自己造成的。\\ 
    C. 指福与祸没有定数，享福太甚就隐伏着祸端。亦作"福为祸先"。\\ 
    D. 佛家用语，命中注定的灾祸。命中注定的灾祸难以逃脱。\\  
    \textcolor{translationcolor}{Question: What is the meaning of the idiom \textquotedblleft 祸福无门，唯人所召\textquotedblright (Fortune and Misfortune Have No Gate; They Are Summoned by One's Own Deeds.) in the sentence? \\ 
    A. Misfortunes recur due to one's lack of virtue. \\ 
    B. Disasters and blessings are not predetermined but caused by people's own actions. \\ 
    C. Fortune and misfortune are unpredictable; excessive enjoyment may lead to disaster. Also called "fortune precedes misfortune". \\ 
    D. Buddhist terminology: Predestined calamities that are inescapable.} \\
    \textbf{Output:}\\
    B \\
    \textcolor{translationcolor}{}
\end{tcolorbox}
\caption{Example for task: Idiom RC.}
\end{figure*}

\begin{figure*}[t]
\begin{tcolorbox}[width=\textwidth, colframe=blue!50!black, colback=blue!5!white, sharp corners]
    \textbf{Task: TCM Syndrome RC}\\
    \textbf{Instruction:} \\以下是关于中医辨证的单项选择题，请直接给出正确答案的选项。\\
    \textcolor{translationcolor}{The following is a multiple-choice question on Traditional Chinese Medicine (TCM) syndrome differentiation. Provide the correct answer directly.} \\
    \textbf{Input:}\\
    题目: 根据以下病人的情况进行中医辩证。\\
    主诉: 肛内肿物外脱伴肿痛便血1周 \\ 
    描述: 患者1周来反复出现肛内肿物外脱伴肿痛便血，外用药物治疗。大便日行1-2次，质稀不成形，排便畅但伴暗红色大量便血、排便不尽及肛门坠胀感。舌淡红苔白，脉数。无发热恶寒及粘液便。\\ 
    检查: 神清体健，皮肤无异常，腹部平软，双下肢无浮肿。舌脉同上。\\ 
    以下选项中最恰当的中医辨证是？\\ 
    A. 外伤损络证 \\ 
    B. 心脾两虚证 \\ 
    C. 风寒阻络证 \\ 
    D. 湿热下注证 \\
    \textcolor{translationcolor}{Case Presentation: A patient presents with 1-week history of prolapsed hemorrhoids accompanied by swelling, pain, and hematochezia. Stools are loose and frequent (1-2/day) with dark-red blood, incomplete defecation, and anal distension. Tongue: pale-red with white coating; Pulse: rapid. No fever/chills or mucous stool. Physical exam unremarkable.} \\
    \textbf{Output:}\\
    D \\
\end{tcolorbox}
\caption{Example for task: TCM Syndrome RC.}
\end{figure*}

\begin{figure*}[t]
\begin{tcolorbox}[width=\textwidth, 
                  colframe=blue!50!black, 
                  colback=blue!5!white, 
                  sharp corners,
                  fontupper=\normalfont] 
    \textbf{Task: Allegorical Saying QA}\\
    \textbf{Instruction:} \\给出以下歇后语的后半句谜底。\\
    \textcolor{translationcolor}{Complete the second metaphorical half of the Allegorical Saying QA (traditional Chinese riddle idiom).} \\
    \textbf{Input:}\\
    墙头草随风倒\\
    \textcolor{translationcolor}{``The wall-top grass sways with the wind'' --- Symbolizing opportunistic vacillation} \\
    \textbf{Output:}\\
    两面装好人\\
    \textcolor{translationcolor}{``Maintains dual facades of virtue'' --- \textit{Liǎngmiàn zhuāng hǎorén} (lit. two-faced pretense of goodness), satirizing hypocrisy through the pun: 装 (pretend) vs. 伪 (false)}
\end{tcolorbox}
\caption{Example for task: Allegorical Saying QA.}
\end{figure*}

\begin{figure*}[t]
\begin{tcolorbox}[width=\textwidth, colframe=blue!50!black, colback=blue!5!white, sharp corners]
    \textbf{Task: Loan Character QA}\\
    \textbf{Instruction:} \\请帮我找出下面文言文句子中的通假字。\\
    \textcolor{translationcolor}{Please help me identify the loan characters in the following classical Chinese sentence.} \\
    \textbf{Input:}\\
    人之爲言，苟亦無信。\\
    \textcolor{translationcolor}{``People's words, if insincere, should not be trusted.''} \\
    \textbf{Output:}\\
    \textquotedblleft 爲\textquotedblright, 诈伪之言。爲，通\textquotedblleft 僞\textquotedblright。\\
    \textcolor{translationcolor}{\textquotedblleft 爲\textquotedblright\ (wéi), meaning deceitful words. Here, 爲 functions as a loan character for 僞 (wěi, 'false').}
\end{tcolorbox}
\caption{Example for task: Loan Character QA.}
\end{figure*}

\begin{figure*}[t]
\begin{tcolorbox}[width=\textwidth, colframe=blue!50!black, colback=blue!5!white, sharp corners]
\textbf{Task: Book Author} \\
\textbf{Input:}\\
请问《关大王独赴单刀会》的作者是谁？ \\ \textcolor{translationcolor}{Who is the author of "King Guan Attending the Single Sword Meeting Alone"?} \\
\textbf{Output:}\\
关汉卿 \; \textcolor{translationcolor}{Guan Hanqing}
\end{tcolorbox}
\caption{Example for task: Book Author. }
\end{figure*}

\begin{figure*}[t]
\begin{tcolorbox}[width=\textwidth, colframe=blue!50!black, colback=blue!5!white, sharp corners]
\textbf{Task: Book Dynasty}\\
\textbf{Input:}\\
请问《涉异志》这本书写成于哪个朝代？ \\ \textcolor{translationcolor}{In which dynasty was "Records of Encountering Anomalies" written?} \\
\textbf{Output:}\\
明 ; \textcolor{translationcolor}{Ming Dynasty}
\end{tcolorbox}
\caption{Example for task: Book Dynasty. }
\end{figure*}

\begin{figure*}[t]
\begin{tcolorbox}[width=\textwidth, colframe=blue!50!black, colback=blue!5!white, sharp corners]
\textbf{Task: Book Collection Classification}\\
\textbf{Instruction:} \\ 请将提供的书归类到易藏、医藏、艺藏、诗藏、史藏、子藏、集藏、儒藏、道藏佛藏中的一类。\\
\textcolor{translationcolor}{Classify the provided book into one of the following categories: Yi (Changes), Medicine, Arts, Poetry, History, Philosophy, Collections, Confucianism, Taoism and Buddhism. } \\
\textbf{Input:}\\
《伤寒论集注》 \; \textcolor{translationcolor}{Collected Commentary on the Treatise on Cold Damage Disorders} \\
\textbf{Output:}\\
医藏 ; \textcolor{translationcolor}{Medicine Collection}
\end{tcolorbox}
\caption{Example for task: Book Collection Classification. }
\end{figure*}

\begin{figure*}[t]
\begin{tcolorbox}[width=\textwidth, colframe=blue!50!black, colback=blue!5!white, sharp corners]
\textbf{Task: Poetry Generation}\\
\textbf{Instruction:} \\ 给出一首符合下述关键词的五言诗、七言诗或者词。\\
\textcolor{translationcolor}{Generate a classical poem (wuyan/qiyan/ci) that embodies the given thematic keywords.} \\
\textbf{Input:}\\
写雪、抒情 \; \textcolor{translationcolor}{Themes: Snowscape depiction; Lyric expression} \\
\textbf{Output:}\\
雪散因和气，冰开得暖光。\\
春销不得处，唯有鬓边霜。\\
\textcolor{translationcolor}{Snow melts by gentle warmth, ice thaws into radiant light.\\
Spring cannot erase all traces---only frost at the temples remains.\\
(Bai Juyi, Tang Dynasty, "Musing on Transience")\\
\footnotesize{*五言绝句 \textit{wuyan jueju} (five-character quatrain)} \\
\footnotesize{*七言律诗 \textit{qiyan lüshi} (seven-character regulated verse)} \\
\footnotesize{*词 \textit{ci} (Lyrical poetry adhering to specific \textit{cipai} (词牌) tonal patterns.)}}
\end{tcolorbox}
\caption{Example for task: Poetry Generation. }
\end{figure*}

\begin{figure*}[t]
\begin{tcolorbox}[width=\textwidth, colframe=blue!50!black, colback=blue!5!white, sharp corners]
\textbf{Task: Poetry Line Source Tracing}\\
\textbf{Instruction:} \\ 请给出以下句子的出处。\\
\textcolor{translationcolor}{Please provide the source of the following poetic lines. } \\
\textbf{Input:}\\
草枯鹰眼疾, 雪尽马蹄轻。 \\ \textcolor{translationcolor}{Withered grass sharpens eagle eyes, melted snow lightens horse hooves. } \\
\textbf{Output:}\\
王维《观猎》 \; \textcolor{translationcolor}{Wang Wei's "Watching the Hunt"}
\end{tcolorbox}
\caption{Example for task: Poetry Line Source Tracing. }
\end{figure*}

\begin{figure*}[t]
\begin{tcolorbox}[width=\textwidth, colframe=blue!50!black, colback=blue!5!white, sharp corners]
    \textbf{Task: Famous Quote Source Tracing}\\
    \textbf{Instruction:} \\请给出以下句子的出处。\\
    \textcolor{translationcolor}{Please identify the source of the following classical Chinese quotation.} \\
    \textbf{Input:}\\
    \textquotedblleft 尽信书，则不如无书。\textquotedblright\\
    \textcolor{translationcolor}{\textquotedblleft If one believes everything in books, it would be better not to have books at all.\textquotedblright} \\
    \textbf{Output:}\\
    《孟子·尽心章句下·第三节》\\
    \textcolor{translationcolor}{\textit{Mèngzǐ · Jìn Xīn II · Section 3} (Mencius, Jin Xin Chapter Part 2), \\ 
    \footnotesize{*This dictum reflects Mencius' emphasis on critical thinking in Confucian philosophy.}}
\end{tcolorbox}
\caption{Example for task: Famous Quote Source Tracing.}
\end{figure*}

\begin{figure*}[t]
\begin{tcolorbox}[width=\textwidth, colframe=blue!50!black, colback=blue!5!white, sharp corners]
\textbf{Task: Idiom Source Tracing}\\
\textbf{Instruction:} \\ 给出以下成语的出处。\\
\textcolor{translationcolor}{Provide the source of the following idiom. } \\
\textbf{Input:}\\
老马识途 ; \textcolor{translationcolor}{An old horse knows the way} \\
\textbf{Output:}\\
《韩非子·说林上》管仲、隰朋从于桓公伐孤竹, 春往冬返, 迷惑失道。管仲曰'老马之智可用也。'乃放老马而随之。遂得道。\\
\textcolor{translationcolor}{From "Han Feizi · Forest of Persuasion I": When Guan Zhong and Xi Peng followed Duke Huan to attack Guzhu, they lost their way returning in winter after departing in spring. Guan Zhong said, "The wisdom of an old horse can be useful. " They then let the old horse lead the way and found their path. }
\end{tcolorbox}
\caption{Example for task: Idiom Source Tracing. }
\end{figure*}

\begin{figure*}[t]
\begin{tcolorbox}[width=\textwidth, colframe=blue!50!black, colback=blue!5!white, sharp corners]
    \textbf{Task: Inverse Poetry Translation}\\
    \textbf{Instruction:} \\根据以下翻译给出对应的古诗词原文。\\
    \textcolor{translationcolor}{Provide the original classical Chinese poem corresponding to the following modern translation.} \\
    \textbf{Input:}\\
    预备丰盛的饭菜，邀请我到他好客的农家。\\
    翠绿的树林围绕着村落，苍青的山峦在城外横卧。\\
    推开窗户面对谷场菜园，手举酒杯闲谈庄稼情况。\\
    等到九九重阳节到来时，再请君来这里观赏菊花。\\
    \textcolor{translationcolor}{My old friend prepares chicken and millet feast, inviting me to his rustic abode.\\
    Emerald woods embrace the village, azure hills recline beyond the walls.\\
    Opening windows to threshing fields, we toast and speak of crops.\\
    When the Double Ninth Festival arrives, I'll return to admire chrysanthemums.} \\
    \textbf{Output:}\\
    题目: 过故人庄 \; 作者: 孟浩然 \\
    故人具鸡黍，邀我至田家。\\
    绿树村边合，青山郭外斜。\\
    开轩面场圃，把酒话桑麻。\\
    待到重阳日，还来就菊花。\\
    \textcolor{translationcolor}{Title: Passing the Old Friend's Village \; Author: Mèng Hàorán (Meng Haoran, Tang Dynasty poet) \\
    \textquotedblleft My friend has chicken and millet prepared,\\ 
    Bidding me come to his rustic cottage.\\ 
    Green trees encircle the village,\\ 
    Blue hills slant beyond the outer walls.\\ 
    Open the windows facing field and garden,\\ 
    Raise wine cups discussing mulberry and hemp.\\ 
    When comes the Double Ninth Day,\\ 
    Again I'll come to savor chrysanthemums.\textquotedblright \\
    \footnotesize{* \textit{桑麻} (sāng má) symbolizes agricultural affairs; \textit{重阳} (Chóngyáng) refers to the traditional festival on the 9th day of the 9th lunar month.}}
\end{tcolorbox}
\caption{Example for task: Inverse Poetry Translation.}
\end{figure*}

\begin{figure*}[t]
\begin{tcolorbox}[width=\textwidth, colframe=blue!50!black, colback=blue!5!white, sharp corners]
    \textbf{Task: Poetry Translation}\\
    \textbf{Instruction:} \\翻译以下诗词。\\
    \textcolor{translationcolor}{Translate the following classical Chinese poem into modern Chinese prose.} \\
    \textbf{Input:}\\
    题目: 闺怨 \; 作者: 王昌龄 \\
    闺中少妇不知愁，春日凝妆上翠楼。\\
    忽见陌头杨柳色，悔教夫婿觅封侯。\\
    \textcolor{translationcolor}{Title: Boudoir Lament \; Author: Wang Changling \\
    \textquotedblleft The young wife in her chamber knows not sorrow's trace,\\
    Adorned in spring splendor, ascends the jade tower's grace.\\
    Sudden sight of willow hues by the roadside lane,\\
    Regrets urging her lord to seek honors on the frontier plain.\textquotedblright} \\
    \textbf{Output:}\\
    闺中少妇未曾有过相思离别之愁，在明媚的春日，她精心妆饰，登上高楼。\\
    忽然看到路边的杨柳春色，惆怅之情涌上心头。她后悔当初不该让丈夫从军边塞，建功封侯。\\
    \textcolor{translationcolor}{The young gentlewoman, untouched by parting grief in her chamber,\\
    Ascends the emerald tower in radiant spring, adorned with care.\\
    Startled by willow-green hues along the path,\\
    Regret floods her heart---why bid her lord seek glory on distant frontiers? \\
    \footnotesize{* \textit{觅封侯} (mì fēnghóu) refers to pursuing military honors during the Tang frontier campaigns.}}
\end{tcolorbox}
\caption{Example for task: Poetry Translation.}
\end{figure*}

\begin{figure*}[t]
\begin{tcolorbox}[width=\textwidth, colframe=blue!50!black, colback=blue!5!white, sharp corners]
    \textbf{Task: Ancient Chinese Translation}\\
    \textbf{Instruction:} \\将这句话翻译为现代汉语：\\
    \textcolor{translationcolor}{Translate the following classical Chinese sentence into modern Chinese.} \\
    \textbf{Input:}\\
    秋，征堪、向，欲以为谏大夫，恭、显白皆为中郎。\\
    \textcolor{translationcolor}{\textquotedblleft In autumn, [the emperor] summoned Zhou Kan and Liu Xiang, intending to appoint them as Remonstrance Grand Masters. However, Hong Gong and Shi Xian reported [against this], and all were made Central Lang Officers.\textquotedblright} \\
    \textbf{Output:}\\
    秋天，征召周堪、刘向，想任命为谏大夫，弘恭、石显禀告，都做了中郎。\\
    \textcolor{translationcolor}{
    In autumn, Zhou Kan and Liu Xiang were summoned to the court with the intention of appointing them as Remonstrance Grand Masters (Jiàn Dàfū 谏大夫). However, Hong Gong (Hóng Gōng 弘恭) and Shi Xian (Shí Xiǎn 石显), through their memorials (bái 白), intervened, resulting in both scholars being appointed instead as Central Lang Officers (Zhōng Láng 中郎), a rank of palace guards.\\
    \footnotesize{*征堪、向, 周堪(Zhōu Kān)、刘向(Liú Xiàng) Renowned scholars of the Han Dynasty\\}
    \footnotesize{*谏大夫(Jiàn Dàfū), 监察官职 Remonstrance Grand Master\\}
    \footnotesize{*中郎(Zhōng Láng), 宫廷侍卫官 Central Lang Officer\\}
    \footnotesize{*白(Bái), 禀告奏议的缩略语 Abbreviation for reporting memorials\\}
    }
\end{tcolorbox}
\caption{Example for task: Ancient Chinese Translation.}
\end{figure*}

\begin{figure*}[t]
\begin{tcolorbox}[width=\textwidth, colframe=blue!50!black, colback=blue!5!white, sharp corners]
    \textbf{Task: Poetry Appreciation}\\
    \textbf{Instruction:} \\赏析以下诗词。\\
    \textcolor{translationcolor}{Analyze the following classical Chinese poetry.} \\
    \textbf{Input:}\\
    小令尊前见玉箫。银灯一曲太妖娆。歌中醉倒谁能恨，唱罢归来酒未消。\\
    春悄悄，夜迢迢。碧云天共楚宫遥。梦魂惯得无拘检，又踏杨花过谢桥。\\
    \textcolor{translationcolor}{Title: \textit{Partridge Sky} (鹧鸪天) \; Author: Yàn Jǐdào (晏几道, Song Dynasty)}\\
    
    \textbf{Output:}\\
    \begin{minipage}[t]{0.48\textwidth}
        \textbf{上阕解析}\\
        词写春夜怀人，深情款款。起句用韦皋玉箫典故，暗喻前世情缘。\\
        \textcolor{translationcolor}{\footnotesize{The opening couplet alludes to the legendary romance between Wéi Gāo and Jade Xiao, symbolizing predestined love. The "silver lamp" illuminates an enchanting musical performance, showcasing the poet's profound captivation.}}
    \end{minipage}
    \hfill
    \begin{minipage}[t]{0.48\textwidth}
        \textbf{下阕解析}\\
        春夜迢迢，楚宫隐喻美人遥不可及。末句"踏杨花过谢桥"乃千古名句，\\ 
        \textcolor{translationcolor}{\footnotesize{The imagery of "treading catkins past Xie Bridge" (踏杨花过谢桥) embodies transcendent dream-journeys, becoming an iconic representation of unconstrained spiritual longing in Chinese literature.}}
    \end{minipage}\\
    
    \textbf{艺术特色}\\
         虚实相生：宴饮之乐与孤眠清寂的强烈对比\\
         用典技法：玉箫典故深化情感厚度\\
         时空处理：现实物理时空 vs. 梦境心理时空\\
    \textcolor{translationcolor}{\\
        \textit{Duality of presence}: Contrast between banquet revelry and solitary longing\\
        \textit{Intertextuality}: Historical allusion enhances emotional depth\\
        \textit{Chronotopic innovation}: Physical-temporal constraints vs. dreamscape freedom}\\
    \footnotesize{
        * 韦皋玉箫故事见《云溪友议》卷三\\
        ** 谢桥意象源于谢娘典故，象征才子佳人相会之地\\
        \textcolor{translationcolor}{* The Wei-Jade legend from \textit{Records of Yunxi Friends} (云溪友议) \\
        ** Xie Bridge symbolizes literati-romantic encounters in classical tradition}
    }
\end{tcolorbox}
\caption{Example for task: Poetry Appreciation.}
\end{figure*}

\begin{figure*}[t]
\begin{tcolorbox}[width=\textwidth, colframe=blue!50!black, colback=blue!5!white, sharp corners]
\textbf{Task: Idiom Explanation}\\
\textbf{Instruction:} \\ 给出以下成语的解释。\\
\textcolor{translationcolor}{Provide the explanation for the following idiom. } \\
\textbf{Input:}\\
衅起萧墙 ; \textcolor{translationcolor}{Trouble begins within the inner walls} \\
\textbf{Output:}\\
萧墙古代宫室内当门的小墙。指祸乱发生在家里。比喻内部发生祸乱。\\
\textcolor{translationcolor}{Xiaowang refers to the small wall inside ancient palace doors. It indicates trouble arising within one's own household, metaphorically describing internal strife. }
\end{tcolorbox}
\caption{Example for task: Idiom Explanation. }
\end{figure*}

\begin{figure*}[t]
\begin{tcolorbox}[width=\textwidth, colframe=blue!50!black, colback=blue!5!white, sharp corners]
\textbf{Task: TCM QA}\\
\textbf{Instruction:} \\ 根据以下文本回答问题。\\
\textcolor{translationcolor}{Answer the question based on the following text.} \\
\textbf{Input:}\\
饮食减肥是很多人在春天减肥最常见的一种方法。不仅是因为饮食减肥效果最显著，还因为饮食减肥方法最简单。今天，要为大家推荐的就是减肥水果和汤粥，快来看看都有哪些吧。\\
巧选水果。\\
选择含糖较少的水果。据研究，菠萝、哈蜜瓜、木瓜、奇异果、香蕉、葡萄等水果的血糖指数较高，减肥族应避免摄取太多这类水果。而像苹果、猕猴桃、柠檬、李子、樱桃、柑橘类等血糖指数较低，是减肥族在搭配水果餐时的较佳选择。\\
餐前吃水果。\\
想要水果帮助我们实现减肥的效果，您在生活中就要注意进行水果饮食的时间。研究表明：如在进餐前20分钟至40分钟吃一些水果或饮用1至2杯果汁，则可防止进餐过多导致的肥胖。因为水果或果汁中富含果糖和葡萄糖，可快速被机体吸收，提高血糖浓度，降低食欲。水果内的粗纤维还可让胃部有饱胀感。合理使用水果。饮食对于我们健康的重要性是不能忽视的，尤其是水果的饮食，您就要更加的注意。水果的食用量不要过多。大多数人认为，水果富含纤维素，几乎不含脂肪和蛋白质，因而可以无节制地放心食用，其实这是一个误区。水果并非能量很低的食品，由于味道甜美很容易吃得过多，其中的糖就会转化为脂肪而堆积。\\

问题：减肥适合吃哪几种水果？\\

\textcolor{translationcolor}{Dietary regulation remains the most prevalent weight-loss approach in spring, valued for its efficacy and simplicity. This article recommends fruits and soups for weight management, with evidence-based strategies:\\
Strategic Fruit Selection\\
Prioritize low-glycemic-index fruits: Apples, kiwifruit, lemons, plums, cherries, and citrus varieties.
Avoid high-glycemic fruits such as pineapple, honeydew melon, papaya, banana, and grapes, as indicated by glycemic load studies.\\
Temporal Optimization\\
Consume fruits or 100\% fruit juice 20-40 minutes before meals. This timing:
Facilitates rapid absorption of fructose and glucose, which elevates blood glucose levels and suppresses appetite.
Leverages dietary fiber to induce gastric distension, thereby enhancing satiety signals.
Quantitative Control
Counteract the common misconception that fruits can be consumed without limits:
Fruits contain non-negligible caloric density (4 kcal per gram of carbohydrates).
Excessive intake triggers hepatic lipogenesis, converting sugars into stored fat.
}

\textbf{Output:}\\
苹果、猕猴桃、柠檬、李子、樱桃、柑橘类等血糖指数较低，是减肥族在搭配水果餐时的较佳选择。\\
\textcolor{translationcolor}{Apples, kiwis, lemons, plums, cherries, and citrus fruits with low glycemic index are optimal choices for weight-loss diets.}
\end{tcolorbox}
\caption{Example for task: TCM QA.}
\end{figure*}

\begin{figure*}[t]
\begin{tcolorbox}[width=\textwidth, colframe=blue!50!black, colback=blue!5!white, sharp corners]
\textbf{Task: Prescription Explanation}\\
\textbf{Instruction:} \\ 给出以下中药方剂的成份以及功效。\\
\textcolor{translationcolor}{Provide the ingredients and efficacy of the following Chinese medicine prescription. } \\
\textbf{Input:}\\
竹叶汤 \; \textcolor{translationcolor}{Bamboo Leaf Decoction} \\
\textbf{Output:}\\
成份: 淡竹叶、黄芩、升麻、木通、车前子、黄连、玄参、芒消、栀子、大黄（炒）。\\
功效: 肝脏实热, 眼赤肿痛。\\
\textcolor{translationcolor}{Ingredients: Lophatherum gracile, Scutellaria baicalensis, Cimicifuga heracleifolia, Akebia quinata, Plantago asiatica, Coptis chinensis, Scrophularia ningpoensis, Natrii Sulfas, Gardenia jasminoides, processed Rheum officinale. \\
Efficacy: Treats liver heat and red, swollen, painful eyes. }
\end{tcolorbox}
\caption{Example for task: Prescription Explanation. }
\end{figure*}

\begin{figure*}[t]
\begin{tcolorbox}[width=\textwidth, colframe=blue!50!black, colback=blue!5!white, sharp corners]
\textbf{Task: Couplet Generation}\\
\textbf{Instruction:} \\ 对对联, 请根据上联生成下联。\\
\textcolor{translationcolor}{Generate the second line of a couplet based on the first line. } \\
\textbf{Input:}\\
荷出污泥而不染, 品格高超可为友 \\
\textcolor{translationcolor}{The lotus emerges from the mud unstained; its noble character makes it worthy of friendship} \\
\textbf{Output:}\\
竹生有节且虚心, 性质坚韧能抵风 \\
\textcolor{translationcolor}{The bamboo grows with joints and humble heart; its resilient nature can resist the wind}
\end{tcolorbox}
\caption{Example for task: Couplet Generation. }
\end{figure*}

\begin{figure*}[t]
\begin{tcolorbox}[width=\textwidth, colframe=blue!50!black, colback=blue!5!white, sharp corners]
    \textbf{Task: Ci Generation}\\
    \textbf{Instruction:} \\按照提供的词牌名和题目写一首词, 要求严格遵守词牌的格律。\\
    \textcolor{translationcolor}{Write a Ci poem according to the provided Cipai (词牌, Ci Tune Pattern) and title, strictly following the metrical rules of the Cipai. } \\
    \textbf{Input:}\\
    词牌: 望江南 \;  \textcolor{translationcolor}{Cipai: Gazing at the Yangtze's South} \\
    题目: 红桥春游词 \; \textcolor{translationcolor}{Title: A Spring Outing at Hong Bridge} \\ 
    \textbf{Output:}\\
    江南好, 修袅袅溪光。杨柳风微莺嘴滑, 桃花漾足鸭头香。几处洗衣裳。 \\
    \textcolor{translationcolor}{The South of Yangtze is lovely, \\
    I seize the creek's gleam for the spring ritual. \\
    In gentle breeze willows sway, while warblers' tongues slip sweet songs, \\
    Peach blossoms flood the scene with fragrance thick as ducks' heads, \\
    Here and there, women wash their clothes. }
\end{tcolorbox}
\caption{Example for task: Ci Generation. }
\end{figure*}

\end{CJK}
\end{document}